\documentclass[11pt,oneside,a4paper,onecolumn]{elsarticle}

\usepackage{supertabular}
\usepackage{longtable,tabularx}
\usepackage{multirow}
\usepackage{times}
\usepackage{fullpage}

\usepackage{gensymb}
\usepackage{tikz}
\usepackage{tikzscale}
\usepackage{alltt}
\usepackage{upquote}
\usepackage{float}
\usepackage{makecell}
\usepackage{array}
\usetikzlibrary{positioning}
\usepackage{graphicx}
\usepackage[toc,page]{appendix}
\usepackage[utf8]{inputenc}
\usepackage{amsmath,mathtools}
\usetikzlibrary{calc,decorations.markings,fit,shapes,chains,arrows}
\usepackage{amsthm}
\usepackage{amsfonts}
\usepackage{amssymb}
\usepackage{enumitem}
\usepackage{caption}
\usepackage{subcaption}
\usepackage{scalerel}
\usepackage{algorithm}
\usepackage{algorithmic}

\usepackage{multicol}

\usepackage{multirow}
\usepackage{url}
\usepackage{booktabs}
\usepackage{nicefrac}
\newcommand{\pder}[2][]{\frac{\partial#1}{\partial#2}}
\newcommand{\der}[2][]{\frac{d#1}{d#2}}

\usepackage{soul}
\usepackage{slashed}
\usepackage{lineno}

\usepackage{xcolor}
\modulolinenumbers[5]

\usepackage[final]{hyperref} % adds hyper links inside the generated pdf file
\hypersetup{
	colorlinks=true,       % false: boxed links; true: colored links
	linkcolor=blue,        % color of internal links
	citecolor=blue,        % color of links to bibliography
	filecolor=magenta,     % color of file links
	urlcolor=blue         
}

\newcolumntype{M}[1]{>{\centering\arraybackslash}m{#1}}
\newcolumntype{N}{@{}m{0pt}@{}}

\tikzset{>=latex}
\def\@author#1{\g@addto@macro\elsauthors{\normalsize%
    \def\baselinestretch{1}%
    \upshape\authorsep#1\unskip\textsuperscript{%
      \ifx\@fnmark\@empty\else\unskip\sep\@fnmark\let\sep=,\fi
      \ifx\@corref\@empty\else\unskip\sep\@corref\let\sep=,\fi
      }%
    \def\authorsep{\unskip,\space}%
    \global\let\@fnmark\@empty
    \global\let\@corref\@empty
    \global\let\sep\@empty}%
    \@eadauthor={#1}
}

\graphicspath{{Images/}}

\begin{document}
\begin{frontmatter}
\title{Predicting Wave Dynamics using Deep Learning with Multistep Integration Inspired Attention and Physics-Based Loss Decomposition}

\author[ubc]{Indu Kant Deo\corref{cor1}}
\ead{indukant@mail.ubc.ca}

\author[ubc]{Rajeev K. Jaiman}
\ead{rjaiman@mail.ubc.ca}
\cortext[cor1]{Corresponding author}
\address[ubc]{Department of Mechanical Engineering, The University of British Columbia, Vancouver, BC V6T 1Z4}

%############################ ABSTRACT #################################
\begin{abstract}
In this paper, we present a physics-based deep learning framework for data-driven prediction of wave propagation in fluid media. The proposed approach, termed Multistep Integration-Inspired Attention (MI2A), combines a denoising-based convolutional autoencoder for reduced latent representation with an attention-based recurrent neural network with long-short-term memory cells for time evolution of reduced coordinates.
This proposed architecture draws inspiration from classical linear multistep methods to enhance stability and long-horizon accuracy in latent-time integration. Despite the efficiency of hybrid neural architectures in modeling wave dynamics, autoregressive predictions are often prone to accumulating phase and amplitude errors over time. 
To mitigate this issue within the MI2A framework, we introduce a novel loss decomposition strategy that explicitly separates the training loss function into distinct phase and amplitude components.
We assess the performance of MI2A against two baseline reduced-order models trained with standard mean-squared error loss: a sequence-to-sequence recurrent neural network and a variant using Luong-style attention.
To demonstrate the effectiveness of the MI2A model, we consider three benchmark wave propagation problems of increasing complexity, namely one-dimensional linear convection, the nonlinear viscous Burgers equation, and the two-dimensional Saint-Venant shallow water system. 
Our results demonstrate that the MI2A framework significantly improves the accuracy and stability of long-term predictions, accurately preserving wave amplitude and phase characteristics. Compared to the standard long-short term memory and attention-based models, MI2A-based deep learning exhibits superior generalization and temporal accuracy, making it a promising tool for real-time wave modeling.

\smallskip
\smallskip

\textbf{Keywords.} Attention-mechanism, Neural Networks, Deep Learning, Reduced-order Models, Numerical Analysis, Multi-step Methods, Wave Propagation, Underwater Acoustics  \end{abstract}
\end{frontmatter}

% \begin{linenumbers}

%############################ INTRO #################################
\section{Introduction}
The study of dynamical systems is central to understanding a broad spectrum of scientific and engineering phenomena, including climate modeling, wave dynamics, and underwater acoustics. In oceanic environments, acoustic noise generated by marine vessels poses a significant threat to marine ecosystems, necessitating predictive models that can inform effective noise mitigation strategies \cite{duarte2021soundscape}.Developing effective noise reduction strategies requires fast and accurate models capable of real-time prediction \cite{VENKATESHWARAN2024118687, deo2024continual, deo2024predicting}. Hyperbolic partial differential equations (PDEs) are commonly employed to model wave propagation in these settings \cite{leveque2002finite}. Solving these equations with high accuracy enables improved noise prediction and control strategies. Traditional numerical approaches, such as finite-difference methods, finite-volume methods, and multistep time integration schemes \cite{alterman1968propagation, leveque2002finite, Iserles_2008}, provide mathematically rigorous frameworks for modeling these systems. While these methods offer high accuracy, they are computationally expensive due to the high dimensionality of discretized equations, making them impractical for real-time applications \cite{deo2022predicting, deo2023combined, han2018solving}. The increasing availability of high-fidelity simulation data presents an opportunity to explore alternative approaches that can achieve comparable accuracy with significantly reduced computational cost.

% ROMs -> Attention
To overcome the computational bottlenecks of traditional solvers, reduced-order models (ROMs) have emerged as efficient alternatives that approximate full-order systems while preserving essential physical features \cite{quarteroni2014reduced, schilders2008model, deo2024data}. Recent advancements in data-driven methodologies have enabled the development of ROMs that leverage machine learning techniques to enhance computational efficiency without sacrificing accuracy. 
Deep learning models, particularly convolutional and recurrent architectures \cite{miyanawala2019hybrid,gupta2022hybrid, Mallik2024DeepAcoustics, deo2023combined, gao2024finite}, have demonstrated remarkable success in capturing complex spatial and temporal dependencies, making them well suited for modeling wave propagation dynamics. Among these approaches, attention mechanisms \cite{deo2022predicting} have emerged as powerful tools to learn long-range dependencies in sequential data.  Originally developed for natural language processing \cite{bahdanau2014neural, luong2015effective, vaswani2017attention}, attention mechanisms have proven to be effective in time series analysis and dynamical system modeling \cite{niu2021review}. 
By adaptively weighting relevant information, attention mechanisms provide a structured means of capturing spatial and temporal correlations in the wave propagation task, making them a promising candidate for developing efficient real-time ROMs. As a result, integrating attention mechanisms with deep learning-based ROMs presents a promising and computationally efficient framework for surrogate modeling of wave propagation.

% The idea and justification
Building on these developments, our research aims to bridge the gap between classical numerical methods and modern deep learning techniques by developing a data-driven framework that integrates the formulation of multistep time-stepping schemes within the attention architecture \cite{hornik1989multilayer, lin2018resnet}. 
Classical linear multistep schemes such as Adams–Bashforth or backward differentiation methods predict future states by forming linear combinations of several past time levels, thus leveraging temporal redundancy to improve stability and accuracy. In parallel, attention mechanisms identify and weigh relevant past states dynamically, enabling the model to focus adaptively on key dependencies across time.  While these attention methods improve stability and adaptability, they do not fully address long-term prediction challenges in neural networks.
One of the primary challenges in long-term predictions is the accumulation of phase and amplitude errors, which can progressively distort predictions over time \cite{LeCun2015,Goodfellow2016,karpatne2017theory}. 
This issue is particularly pronounced in autoregressive models, a class of models that predict future values based on past observations \cite{Hamilton1994, BoxJenkins1970}. 
Due to their iterative nature, even minor errors in early predictions can propagate, leading to significant deviations from the true dynamics \cite{Rottmann2018,beucler2021enforcing}.

% Physics-based loss function
 Traditional loss functions such as mean squared error (MSE) do not account for distinct error types, treating all deviations uniformly. 
 However, in the context of wave propagation, numerical analysis has long recognized that dissipation (amplitude decay) and dispersion (phase shift) are the primary contributors to solution degradation over time \cite{Kreiss1972, deo2024harnessing}. This insight provides a physics-based structured framework for improving the training objectives of deep learning models applied to wave-like systems.
Recent work by Guen et al. \cite{le2019shape} has shown that decomposing the loss function into interpretable components can improve stability and robustness in time-series forecasting. Inspired by this finding, our work explicitly incorporates numerical error decomposition into the learning process by targeting phase and amplitude errors separately during training.
By integrating numerical error decomposition into machine learning techniques, we propose a physics-based loss decomposition technique to improve predictive accuracy over extended horizons.

% Proposed framework
In this paper, we propose a novel deep learning framework, termed multistep integration-inspired attention (MI2A), which combines the principles of classical numerical time integration with modern attention-based sequence modeling to learn the evolution of nonlinear dynamical systems from data. Our approach builds on the structure of linear multistep methods and extends this concept by replacing fixed integration coefficients with adaptive attention weights that are learned from the data. 
Notably, MI2A is more than just a temporal attention mechanism and can be considered as a neural generalization of classical time integrators, wherein the attention architecture is interpreted as a nonlinear, data-driven analog of multistep time-marching schemes. By linking the architecture with numerical integration, we improve both the interpretability and the temporal stability of the predictions.
To further increase predictive performance, we integrate a physics-based loss decomposition strategy that explicitly targets dispersion and dissipation errors. 
Together, these two formulations yield a hybrid modeling paradigm that is both data-efficient and physics-based, capable of making stable, accurate predictions of wave dynamics for the 1D convection and 2D shallow-water benchmark problems. The MI2A framework provides a foundation for advancing reduced-order modeling in fluid dynamics, acoustics, and other time-dependent physical systems.

% Organization
The remainder of the paper is organized as follows. Section 2 presents the mathematical formulation of our reduced-order modeling framework. Section 3 describes the proposed methodology, which brings together the linear multistep interpretation of MI2A, the convolutional autoencoder-based architecture, multistep time evolution using MI2A, and the complete integration of attention-based updates and learnable derivative approximations. Section 4 describes the data preparation process, including the construction of snapshot matrices for sequence-to-sequence learning and the training strategy. Section 5 presents numerical results for three test problems, such as linear convection, viscous Burgers' equations, and 2D shallow-water wave propagation, and compares our approach with existing deep learning-based reduced-order models. Finally, Section 6 summarizes the main contributions and discusses broader implications and future research directions.

\section{Mathematical Background}
\label{sec:problem_formulation}
% Indu:  Eliminate unncessary adjectives
We begin by establishing the mathematical formulation for parametric time-dependent wave propagation systems. Let $\Omega \subset \mathbb{R}^d$ ($d = 1, 2, 3$) denote the spatial domain, and $\mathcal{M} \subset \mathbb{R}^m$ denote a compact parameter space that governs the physical or geometric properties of the problem. 
A general formulation of a time-dependent parametric partial differential equation can be expressed in an abstract form:
\begin{equation}
\begin{aligned}
\pder{t}{\mathbf{U}(\mathbf{X}, t; \mathbf{\mu})}&=\mathcal{F}\left(\mathbf{U}(\mathbf{X}, t; \mathbf{\mu}), \mathbf{X}, t; \mathbf{\mu}\right)\quad &&\mathrm{on} \quad \Omega \times [0, T] \times \mathcal{M},\\
\mathbf{U}(\mathbf{X},0 ; \mathbf{\mu})&=\mathbf{U}_{0}(\mathbf{\mathbf{X},\mu})\quad &&\mathrm{on} \quad \Omega \times \mathcal{M},\\
\mathbf{U}(\mathbf{X},t ; \mathbf{\mu})&=\mathbf{U}_{\partial\Omega}(\mathbf{X},t,\mu)\quad &&\text{on} \quad {\partial\Omega} \times [0, T] \times \mathcal{M},
\label{eqn:PDE}
\end{aligned}
\end{equation}
where $\Omega \subset \mathbb{R}^{i} \ (i = 1, 2, 3)$ denotes the
spatial domain,  $\mathcal{M} \subset \mathbb{R}^{m}$ is a space of possible physical parameters for the problem, and $\mathcal{F}$ is a generic nonlinear operator describing the dynamics of the system. 
The solution field of the system is represented by $\mathbf{U}$: $\Omega \times [0, T] \times \mathcal{M} \rightarrow \mathbb{R}$ and appropriately chosen initial $\mathbf{U}_{0}(\mathbf{\mathbf{X},\mu})$ and boundary conditions $\mathbf{U}_{\partial\Omega}(\mathbf{X},{t,\mu})$. 
The parameter $\mu$ controls aspects of the problem, such as wave speed and Reynolds number.

Upon applying a numerical discretization in space (e.g., finite volume or finite difference), the continuous PDE system is transformed into a system of ordinary differential equations (ODEs): 
% Semi-discretized ODE
\begin{equation}
\begin{aligned}
\der{t}{\mathbf{U_{N}}(\mathbf{X_{N}}, t; \mathbf{\mu})}&=\mathcal{F}_{N}\left(\mathbf{U_{N}}(\mathbf{X_{N}}, t; \mathbf{\mu}), \mathbf{X_{N}}, t; \mathbf{\mu}\right) &&\mathrm{on}\ \Omega_N \times [0, T] \times \mathcal{M},\\
\mathbf{U_{N}}(\mathbf{X_{N}},0 ; \mathbf{\mu})&=\mathbf{U}_{0}(\mathbf{\mathbf{X_{N}},\mu}) &&\text{on}\ \Omega_N \times \mathcal{M},\\
\mathbf{U_{N}}(\mathbf{X_{N}},t ; \mathbf{\mu})&=\mathbf{U}_{\partial\Omega}(\mathbf{X_{N}},{t,\mu}) &&\mathrm{on}\ {\partial\Omega_N} \times [0, T] \times \mathcal{M},
\end{aligned}
\label{eqn:discrete ODE}
\end{equation}
where $\Omega_N \subset \mathbb{R}^{N}$ is the discretized spatial domain,  $\mathbf{U_{N}}: \Omega_N \times [0, T] \times \mathcal{M} \rightarrow \mathbb{R}^{N}$ is a discrete solution and $N$ is the number of spatial degrees of freedom.
Once spatial discretization is performed, classical time-stepping methods are employed to advance the solution in time \cite{acton2020numerical, sethian1999fast}. 
This results in the following time-integrated representation:
% time integration solution
\begin{equation}
\begin{aligned}
\mathbf{U_{N,{\mu}_{i}}^{({N_{T+1}})}} = \mathcal{G} \left(\mathbf{U_{N,{\mu}_{i}}^{({N_{T}})}}, \mathbf{X_{N}}, N_T; \mathbf{\mu_i}\right), 
\end{aligned}
\label{eqn:discrete ODE dyanmics}
\end{equation}
where $\mathbf{U_{N,{\mu}_{i}}^{({N_{T+1}})}} \in \mathbb{R}^{N}$ is a solution at time step $N_{T+1}$.
% Solution Manifold of parametric PDE
For given $(t ; \mathbf{\mu})$ varying in $[0, T] \times \mathcal{M}$, the set of solution fields of Eq. (\ref{eqn:PDE}) is known  as solution manifold \cite{budd1999geometric}  represented by $\mathbf{S_{U}}$  as:
\begin{align}
\mathbf{S}_{\mathbf{U}}=\left[\mathbf{U}_{N,\mu_{1}}^{(t_{1})}, \ldots, \mathbf{U}_{N,{\mu}_{1}}^{({N_{T}})}, \ldots \ldots , \mathbf{U}_{N, {\mu}_{N_{\mathrm {train }}}}^{(t_{1})}, \ldots, \mathbf{U}_{N, {\mu}_{N_{\mathrm {train }}}}^{(N_{T})}\right].
\end{align}
In practice, the solution manifold $\mathcal{S}_{\mathbf{U}}$ often resides on a low-dimensional, nonlinear subspace of $\mathbb{R}^N$ due to the smooth parametric and temporal dependence of wave solutions.
When $\mathbf{\mu} \in \mathcal{M}$, the solution field of Eq. (\ref{eqn:PDE}) admits a solution for each $t \in[0, T]$. 
The intrinsic dimension of the solution field lying in the solution manifold is at most $n_{\mathbf{\mu}}+1$ \cite{mojgani2021low}, where $n_{\mathbf{\mu}}$ is the number of parameters. 
Instead of solving Eq. (\ref{eqn:PDE}) directly, we seek to construct a reduced-order model that approximates the full solution manifold while significantly reducing computational cost. The next section introduces the methodology for achieving this by learning an efficient reduced representation of the solution space.

\section{Methodology \label{sec:method}}
\begin{figure*}[ht]
    \centering
    \includegraphics[width=\textwidth]{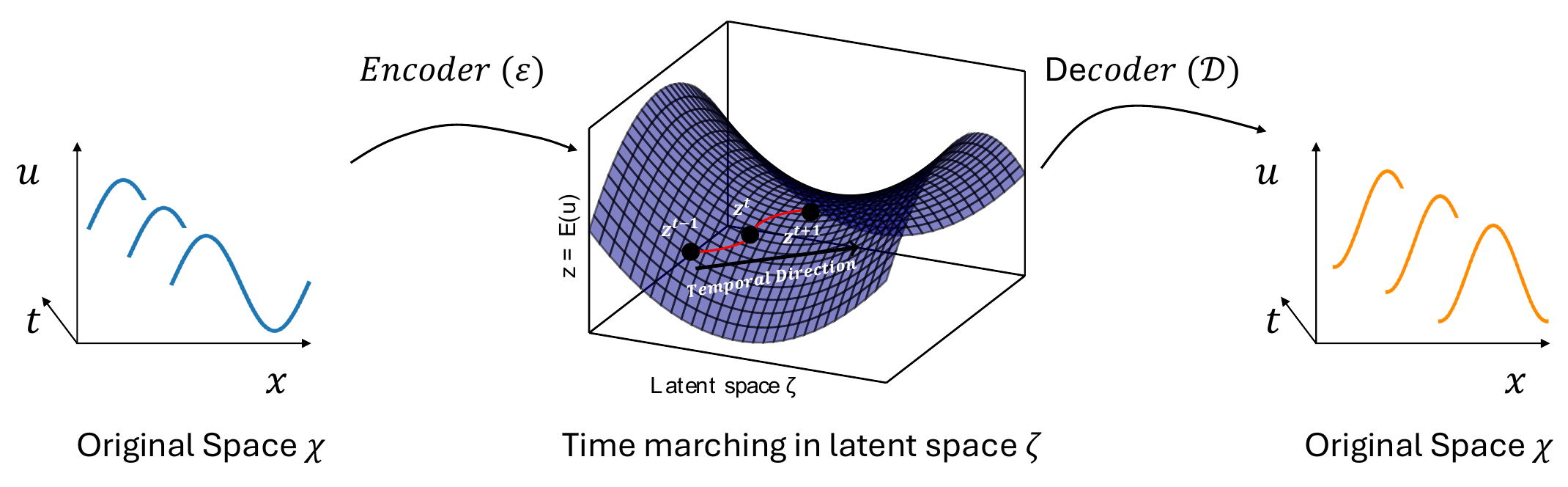}
    \caption{Schematic representation of the encode-propagate-decode architecture, highlighting its key components and their interactions.}
\label{fig:RC-CAN_ROM}
\end{figure*}

In this section, we present the methodology for the proposed Multistep Integration-Inspired Attention (MI2A) framework. MI2A integrates classical time-stepping concepts with modern attention-based sequence modeling to enable efficient and accurate reduced-order prediction of wave dynamics. 
 The MI2A framework consists of three primary components: (1) a convolutional encoder for spatial dimensionality reduction, (2) a recurrent neural network with an attention mechanism to capture temporal dependencies while incorporating multistep integration-inspired dynamics, and (3) a decoder that reconstructs the full-order solution from the latent space. By integrating these components, MI2A efficiently models nonlinear dynamical systems while maintaining numerical stability. Figure \ref{fig:RC-CAN_ROM} illustrates the architecture and its three core components.

\subsection{Convolutional Encoder-Based Dimensionality Reduction}
\begin{figure*}
    \centering
    \includegraphics[width=\textwidth]{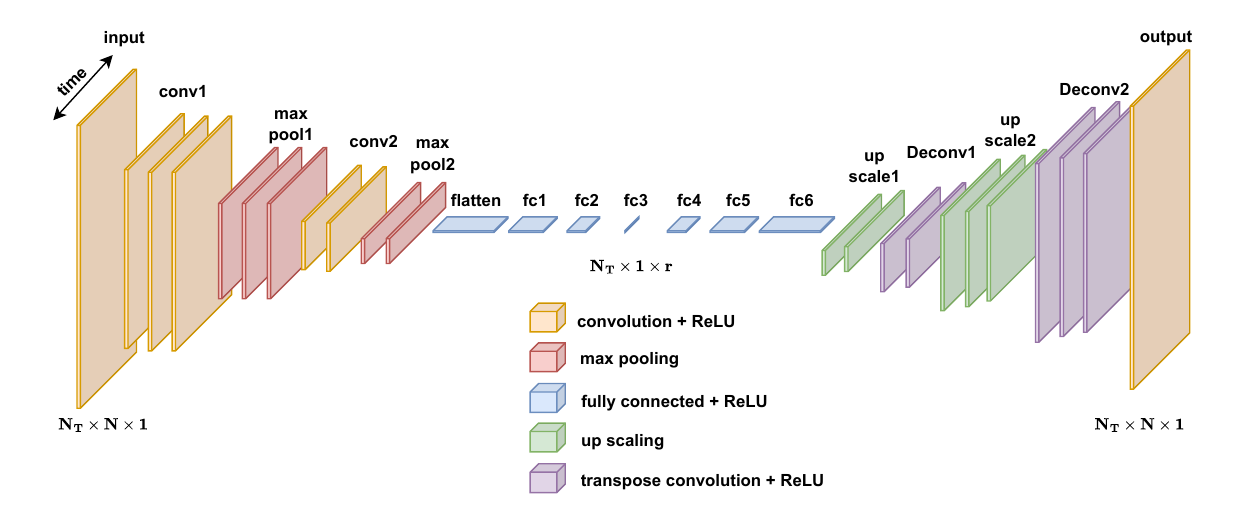}
    \caption{A diagram depicting convolutional autoencoder architecture employed for the dimensionality reduction.}
\label{fig:convAutoLin}
\end{figure*}

We first describe the encoder architecture used to construct a low-dimensional latent representation of the solution snapshots. To efficiently handle high-dimensional solution spaces of PDE, we employ a convolutional encoder \cite{hinton2006reducing, vincent2008extracting, 10.1007/978-3-642-21735-7_7} that maps the full-order spatial state $\mathbf{U}_{N}^{(t)}$ into a lower-dimensional latent representation $\mathbf{Z}^{(t)}$:
\begin{equation}
\mathbf{Z}^{(t)} = f_\theta(\mathbf{U}^{(t)}_N), 
\end{equation}
where $f_\theta$ is the encoder parameterized by neural network weights. $\mathbf{U}^{(t)}_N\in \mathbb{R}^N$ is solution in the physical space and $\mathbf{Z}^{(t)} \in \mathbb{R}^r$ is a solution in the latent space.
The encoder uses convolutional layers to capture spatial structures followed by fully-connected neural network layers to project into a compact latent space. 
This dimensionality-reduction facilitates more effective and computationally efficient modeling of the temporal evolution of high-dimensional dynamical systems, illustrated in Figure~\ref{fig:convAutoLin}.  Notably, convolution-based dimensionality reduction techniques have been shown to be particularly effective in addressing problems with large Kolmogorov n-width \cite{greif2019decay}, such as wave propagation, by projecting them to a non-linear reduced manifold \cite{deo2022predicting, lee2020model, gonzalez2018deep, sorteberg2018approximating, fresca2021comprehensive, maulik2021reduced, XU2020113379}.

\subsection{Linear Multistep Method and its Connection to MI2A}
Once the high-dimensional state $\mathbf{U}^{(t)}_N$ is projected onto the latent representation $\mathbf{Z}^{(t)}$, the latent representation can evolve along the temporal dimension. In classical numerical analysis, linear multistep methods \cite{Iserles_2008} is one of the methods used to calculate the time advancement of semi-descrite ordinary differential equations. The linear multistep method evaluates the next time step by linearly combining past solution states. A general explicit $k$-step linear multistep method is expressed as:
\begin{equation}
\begin{aligned}
&\mathbf{U}_N^{(n+1)} = \sum_{i=0}^{k} \alpha_i \mathbf{U}_N^{(n-i)} + \Delta t \sum_{i=0}^{k} \beta_i \mathcal{F}(\mathbf{U}_N^{(n-i)}, t_{n-i}; \mu), \\
\end{aligned}
\label{eqn:LMM}
\end{equation}
where $\mathbf{U}_N^{(n+1)}$ denotes the state at time step $t_{n+1}$, $\mathcal{F}(\mathbf{U}_N,t_N;\mu)$ is the system’s time derivative, $\Delta t$ is the time step, and ${\alpha_i}, {\beta_i}$ are fixed coefficients chosen to ensure stability and accuracy. Intuitively, linear multistep methods act as linear filters, combining past states and derivatives using fixed weights to evolve the current state. This fixed linearity, while computationally efficient, limits the adaptability to nonlinear and state-dependent behaviors inherent in complex dynamical systems.
To address this limitation, our MI2A approach generalizes these methods by replacing fixed coefficients with learned dynamic weighting functions through an attention mechanism. This transition allows the model to adjust its predictions adaptively, significantly enhancing its capability to capture the nonlinear and nonstationary dynamics of complex systems.

\subsection{Multi-step Time Evolution using MI2A}
In MI2A architecture, the temporal evolution of latent reprsentation $\mathbf{Z}^{(t)}$ is modeled using a recurrent LSTM-based encoder-decoder framework. 
Specifically, the encoder captures temporal correlations, while the decoder with attention is employed to dynamically compute future latent states. Formally, let $\mathbf{Z}^{\left[1: N_T\right]}$ denotes the sequence of $N_T$ reduced latent representations. 
We feed these into an LSTM encoder:
$$
\mathbf{H}_{\mathrm{enc}}, h_{\mathrm{enc}}^{t_N}, c_{\mathrm{enc}}^{t_N}=\operatorname{LSTM}_{\mathrm{enc}}\left(\mathbf{Z}^{\left[1: N_T\right]}\right),
$$
where $\mathbf{H}_{\mathrm{enc}}=\left\{h_{\mathrm{enc}}^{t_1}, \ldots, h_{\mathrm{enc}}^{t_{N_T}}\right\}$ is the sequence of hidden encoder states and $\left(h_{\mathrm{enc}}^{t_N}, c_{\mathrm{enc}}^{t_N}\right)$ are the final hidden and cell states at the last timestep $t_N$. These final states encapsulate a compressed global context derived from all preceding latent representations. Since the latent space evolves over time, it is essential to anchor the decoder to a stable reference state. To ensure consistency during decoding, we utilize the final hidden state of the encoder as input in each decoding step, while the decoder itself is initialized with the encoder’s final hidden and cell states. Mathematically, we form an input matrix:
\begin{equation}
\label{eqn:dec_inp}
\mathbf{H}_{\mathrm {dec-inp }}=\left[h_{\mathrm{enc}}^{t_N}, h_{\mathrm{enc}}^{t_N}, \ldots, N_T \mathrm{-times}\right],
\end{equation}
and feed this into the decoder LSTM:
\begin{equation}
\mathbf{S}_{\mathrm{dec}}, s_{\mathrm{dec}}^{t_N}, p_{\mathrm{dec}}^{t_N}=\operatorname{LSTM}_{\mathrm{dec}}\left(\mathbf{H}_{\mathrm{dec}-\mathrm{inp}}, \mathrm { initial\ state }=\left[h_{\mathrm{enc}}^{t_N}, c_{\mathrm{enc}}^{t_N}\right]\right)
\end{equation}
Here, $\mathbf{S}_{\mathrm{dec}}=\left\{s_{\mathrm{dec}}^{t_1}, \ldots, s_{\mathrm{dec}}^{t_{N_T}}\right\}$ is the sequence of hidden decoder states, while $p_{\mathrm {dec }}^{t_i}$ is the corresponding decoder cell state. Repeating $h_{\mathrm {enc }}^{t_N}$ ensures that a robust global context is introduced into each decoding step, preventing the decoder from drifting away from the encoder's summarizing of the input sequence.

To compute a dynamically weighted combination of the encoder's hidden states, we incorporate an attention mechanism that allows the prediction at time step $t$ to attend to a range of past encoder hidden states, rather than relying solely on the final hidden state. Specifically, at each decoding time step $t$, we compute a context vector $q_t$ as a weighted sum of encoder hidden states:
\begin{equation}
q_t=\sum_{i=1}^{N_T} {\mathrm{{Attention}}}(s_{dec}^t,h_{enc}^{t_i}) \cdot h_{\mathrm{enc}}^{t_i},
\end{equation}
where each $ \mathrm{{Attention}}(s_{dec}^t,h_{enc}^{t_i})$ is an attention weight that reflects how relevant the encoder state $h_{\mathrm {enc }}^{t_i}$ is for predicting the next time-step. These weights are data dependent and are calculated by comparing the current state of the decoder $s_{dec}^t$ with the state of the encoder $h_{\mathrm {enc }}^{t_i}$, typically through a learned scoring function such as the multiplicative score $e_{t, i}=s_{dec}^{t\top} \cdot W \cdot h_{\mathrm {enc }}^{t_i}$. These scores are then normalized via softmax:
\begin{equation}
{\mathrm{{Attention}}}(s_{dec}^t,h_{enc}^{t_i})=\frac{\exp \left(e_{t, i}\right)}{\sum_{j=1}^{N_T} \exp \left(e_{t, j}\right)},
\label{eq:attn}
\end{equation}
 so that $\sum_{i=1}^{N_T}{\mathrm{{Attention}}}(s_{dec}^t,h_{enc}^{t_i})=1.$ Intuitively, attention allows the model to focus on the encoder representations that are the most relevant at time $t$, and these weights change dynamically at each step based on the evolving state of the decoder. In contrast to fixed linear schemes, this is an attention-dependent weighting mechanism.
%%%%%%%%%%%%%%%%%%%%%%%%%%%%%%%%%%%%%%%%%%%%%%%%%%%%%%%%%%%%%%%%%%%%%%%%
\subsection{Attention-Based Updates as a Generalization of Linear Multistep Methods}
Furthermore, we draw a conceptual link between linear multistep methods and attention-based update mechanisms, demonstrating that any linear multistep scheme can be interpreted as a specific instance of the attention framework. In addition, we show that attention mechanisms extend linear multistep updates by allowing for nonlinear and data-dependent dynamic formulations.

To illustrate this connection, we consider the one-dimensional linear convection equation as the governing equation for wave propagation phenomena, given by PDE:
\begin{equation}
    \label{eq:conv1d}
    \frac{\partial \mathbf{U}}{\partial t} + \mu\,\frac{\partial \mathbf{U}}{\partial x} = 0,
\end{equation}
defined on a uniform spatial grid $\{x_i\}$ with spacing $\Delta x$ and positive wave speed $\mu > 0$. A first-order upwind discretization of the spatial derivative yields the semi-discrete formulation:
\begin{equation}
    \label{eq:upwind}
    \frac{d \mathbf{U}_N}{dt} = -\frac{\mu}{\Delta x}\left(\mathbf{U}_N - \mathbf{U}_{N-1}\right),
\end{equation}
which results in a system of ordinary differential equations in time.

To advance the solution in time, we employ a general 
$k$-step linear multistep method. Let $\mathbf{U}_N^{(n)} \approx \mathbf{U}_N(t_n)$, where $t_{n+1} = t_n + \Delta t$  denotes the temporal discretization with time step 
$\Delta t$. The general form of the linear $k$-step multistep method is given by Eq. \ref{eqn:LMM}, with the right-hand side function defined as: 
\begin{equation}
    \label{eqn:rhs}
    \mathcal{F}\left(\mathbf{U}_N^{(n-i)}, t_{n-i}; \mu\right) = -\frac{\mu}{\Delta x}\left(\mathbf{U}_N^{(n-i)} - \mathbf{U}_{N-1}^{(n-i)}\right)   
\end{equation}
which is obtained from the upwind discretization of the spatial derivative.
Substituting the above expression into Eq.~\eqref{eqn:LMM} yields a general recurrence relation of the form:
\begin{equation}
    \label{eq:lmm-expanded}
    \mathbf{U}_N^{(n+1)} = \sum_{i=0}^{k} \gamma_i\,\mathbf{U}_N^{(n-i)} + \sum_{i=0}^{k} \delta_i\,\mathbf{U}_{N-1}^{(n-i)},
\end{equation}
where $\gamma_i$ and $\delta_i$ are constant coefficients determined by $\{\alpha_i, \beta_i\}$, $\mu$, $\Delta t$, and $\Delta x$. Equation~\eqref{eq:lmm-expanded} demonstrates that the updated solution is a fixed linear combination of previously computed values at both $x_i$ and $x_{i-1}$.
For example, let us take the two-step Adams–Bashforth scheme given by:
\begin{equation}
  \mathbf{U}_N^{(n+2)}=  \mathbf{U}_N^{(n+1)}+\frac{ \Delta t}{2}\left[3 \mathcal{F}\left(\mathbf{U}_N^{(n+1)}, t_{n+1}\right)-\mathcal{F}\left( \mathbf{U}_N^{(n)}, t_n\right)\right].
  \label{eq:AB2}
\end{equation}
Putting in the specific form of $\mathcal{F}(\cdot)$ (Eq. \ref{eqn:rhs}) into two-steps Adams-Bashforth scheme (Eq. \ref{eq:AB2}), we obtain:
\begin{equation}
\begin{aligned}
\mathbf{U}_N^{(n+2)} & =\mathbf{U}_N^{(n+1)}+\frac{\Delta t}{2}\left[3\left(-\frac{\mu}{\Delta x}\right)\left(\mathbf{U}_N^{(n+1)}-\mathbf{U}_{N-1}^{(n+1)}\right)-\left(-\frac{\mu}{\Delta x}\right)\left(\mathbf{U}_N^{(n)}-\mathbf{U}_{N-1}^{(n)}\right)\right], \\
& =\mathbf{U}_N^{(n+1)}-\frac{3 \mu\Delta t }{2 \Delta x}\left(\mathbf{U}_N^{(n+1)}-\mathbf{U}_{N-1}^{(n+1)}\right)+\frac{\mu\Delta t }{2 \Delta x}\left(\mathbf{U}_N^{(n)}-\mathbf{U}_{N-1}^{(n)}\right),
\\
& =\underbrace{\left[1-\frac{3 \mu \Delta t }{2 \Delta x}\right]}_{\gamma_1} \mathbf{U}_N^{(n+1)}+\underbrace{\frac{3 \mu\Delta t }{2 \Delta x}}_{\delta_1} \mathbf{U}_{N-1}^{(n+1)}+\underbrace{\frac{\mu\Delta t }{2 \Delta x}}_{\gamma_2} \mathbf{U}_N^{(n)}-\underbrace{\frac{\mu\Delta t }{2 \Delta x}}_{\delta_2} \mathbf{U}_{N-1}^{(n)}.
\label{eq:expanded_AB2}
\end{aligned}
\end{equation}
Hence, $\mathbf{U}_N^{(n+2)}$ is a constant-coefficient linear combination of four previous states: $\left\{\mathbf{U}_N^{(n+1)}, \ldots, \mathbf{U}_{N-1}^{(n)}\right\}$.
Crucially, these coefficients $\gamma_1, \gamma_2, \delta_1, \delta_2$ do not depend on $\mathbf{U}_N$ or $t$ beyond the constant parameters.
We can replicate this exactly in an attention mechanism using an attention model that assigns a constant score of $ \gamma_i$ to each past state $h_{enc}^{t_i}$. Here, the context vector $q_t=\sum_{i=1}^{N_T} \gamma_i h_{\mathrm {enc }}^{t_i}$ remains invariant between time steps. 
This setup mirrors a linear $k$-step method, because each past state $h_{\mathrm {enc }}^{t_i}$ is combined with a fixed, data-independent coefficient $\gamma_i$. Consequently, any linear combination used by linear multistep methods can be viewed as a time-invariant attention scheme.

Although attention can replicate LMM by using these fixed scores, it can also go further. 
In a standard attention framework, the attention weight, ${\mathrm{{Attention}}}(s_{dec}^t,h_{enc}^{t_i})$, is computed by a learned scoring function of both the decoder state $s_{dec}^t$ and each encoder state $h_{enc}^{t_i}$ given by Eq. \ref{eq:attn}.
% \begin{equation}
% \alpha_{t, i}=\operatorname{Align}\left(s_t, h_i\right).
% \end{equation}
%
Because this score changes with the decoder's evolving state, the attention weights become state-dependent, introducing nonlinear coupling. 
From a dynamical system perspective, linear multistep updates yield linear difference equations with constant coefficients, whereas attention-based updates result in difference equations with nonlinear, state-dependent coefficients. 
Thus, attention is more expressive than LMM in capturing complex temporal dependencies.

\subsection{Learning Differential Operator using Convolution}
%%%%%%%%%%%%%%%%%%%%%%%%%%%%%%%%%%%%%%%%%%%%%%%%%%%%%%%%%%%%%%%%%%%%%%%%%%%%%%%%%%%%%%%%%%%%%%%%%%%%%%%%%%%%%%%%%%%%%%%%%%%%%%%%%%%%%%%%%%%%%%%%%%%%%%%%%%%%%%%%%%%%%%%%%%%%%%
We have established that attention generalizes linear multistep methods through its ability to produce either constant or state-dependent weighting of the past input states. 
A key aspect of many time-integration schemes is their dependence not only on the values of the past states but also on the derivatives of the state. In classical LMM, these derivatives appear in the term $\mathcal{F}\left(\mathbf{U}_N^{(n-i)}, t_{N-i} ; \mu\right)$.
To this end, we introduce learnable derivative approximations within the latent space of the model. This approach is inspired by the PDE-Net framework \cite{long2018pde}, which utilizes convolutional filters as data-driven approximations of differential operators. 
Building upon this concept, we integrate convolution-based derivative estimation into the MI2A architecture to increase its expressiveness in capturing the underlying dynamics of complex systems.
Instead of relying on hand-crafted finite difference stencils, the MI2A architecture employs a learnable convolutional operator to approximate temporal derivatives directly in the latent space:
\begin{equation} 
\mathbf{d}^{(n+1)}=\sum_{i=0}^{k-1} \mathbf{W}_{\mathrm{conv }}(i) * {h}_{\mathrm{enc}}^{(n-i)}, \quad \mathbf{D}=\operatorname{Conv1D}\left(\mathbf{H}_{\mathrm{enc}}\right), 
\end{equation} 
where $\mathbf{W}_{\mathrm{conv}}(i)$ denotes trainable convolutional kernels applied to the encoded hidden states $\mathbf{h}_{\mathrm{enc}}^{(n-i)}$, and $\mathbf{D} = \{\mathbf{d}^{(n+1)}, \ldots, \mathbf{d}^{(n+N)}\}$ represents the resulting sequence of derivative estimates at prediction time-steps. 
By learning these convolutional weights directly from trajectory data, the network is able to infer the underlying time-evolution of the system and dynamically adapt its derivative approximations to capture complex and nonlinear dynamics.
%%%%%%%%%%%%%%%%%%%%%%%%%%%%%%%%%%%%%%%%%%%%%%%%%%%%%%%%%%%%%%%%%%%%%%%%%%%%%%%%%%%%
\subsection{MI2A Update Equation and Projection to Physical Space} 
\begin{figure*}
    \centering
    \includegraphics[width=\textwidth]{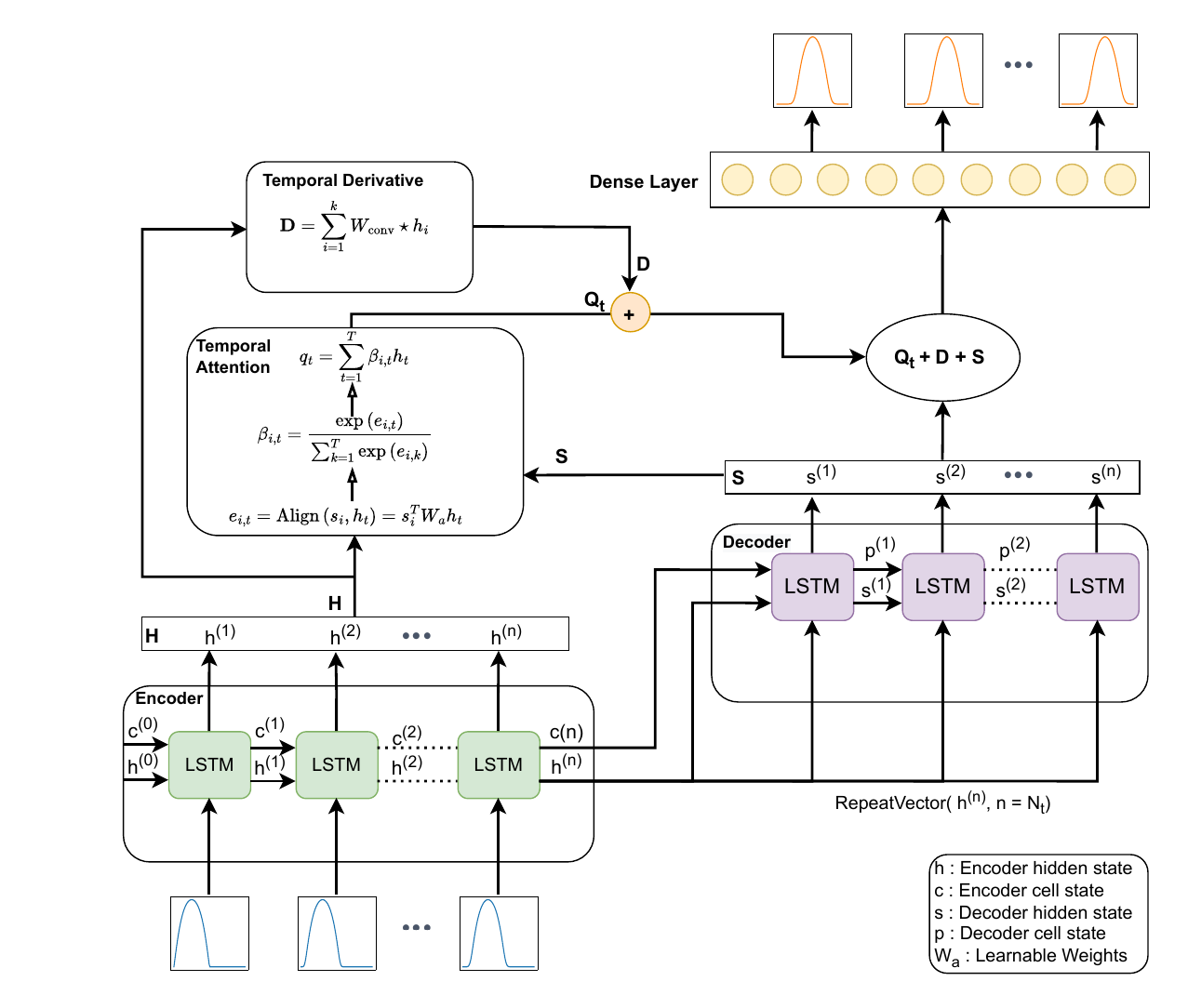}
    \caption{{Illustration of the proposed attention-based sequence-to-sequence evolver. While the encoder generates hidden state vectors $H$ by transforming input, the decoder generates hidden state ($S$) by iterating over final encoder hidden state $h^{(n)}$. Notably the alignment score between $H$ and $S$ are computed.}}
\label{fig:MI2A_prop}
\end{figure*}

Having both the output from attention-based mechanism ($\mathbf{Q}_t$) and the learnable derivative terms ($\mathbf{D}$), we form the final latent state prediction by combining them with the decoder output ($\mathbf{S}_{\mathrm{dec}}$):
\begin{equation}
\mathbf{Z}_{\mathrm{pred}}^{[1:N_T]}=\mathbf{S}_{\mathrm{dec}}+\mathbf{Q}_t+\mathbf{D},
\end{equation}
where $\mathbf{Q}_t =\{q_1,\ldots, q_{t_N}\}$ is the N sequence of outputs from the attention mechanism, and $\mathbf{Z}_{\mathrm{pred}}^{[1:N_T]}$ is the time evolved latent representation.
Together, MI2A unifies these operations to produce a nonlinear generalization of the classical LMM. 
The architecture of MI2A for time marching is shown in Fig. \ref{fig:MI2A_prop}.

\begin{figure*}
    \centering
    \includegraphics[width=\textwidth]{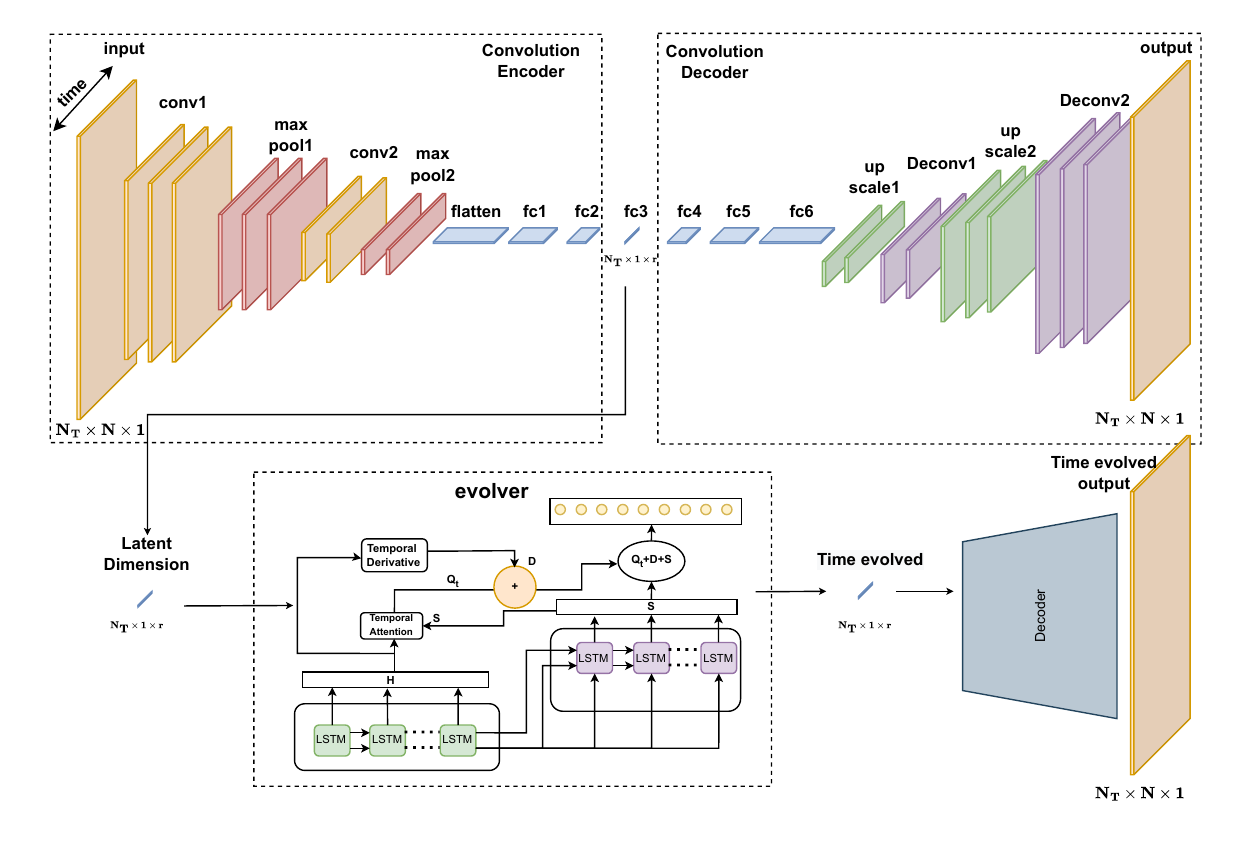}
    \caption{{Visualization of MI2A architecture.  Three blocks are shown namely the convolution encoder for creating the latent low-dimensional representation, the evolver for propagating the low-dimensional feature in time and the decoder for transforming the low dimension space to input data space.}}
\label{fig:MI2A_arch}
\end{figure*}

After evolving the latent state $\mathbf{Z}^{(t)}$ forward in time, the final step is to project the predicted future state in the full-order space. Formally, we apply a learnable decoder 
$g_\phi$ to the latent prediction $\mathbf{Z}_{\mathrm{pred}}^{(t)}$:
\begin{equation}
\mathbf{U}_{\mathrm{pred}}^{(t)} = g_\phi(\mathbf{Z}_{\mathrm{pred}}^{(t)}).
\end{equation}
In other words, $g_\phi$ reverses the dimensionality reduction performed by the encoder, transforming low-dimensional latent features back into the full resolution of the input data space. 
By preserving a learnable mapping in both directions: encoder to compress the high-dimensional input and decoder to reconstruct the predicted output, MI2A can forecast the future states of complex dynamical systems while working primarily in a latent space that is easier to handle computationally.
Figure \ref{fig:MI2A_arch} illustrates the three main components of the MI2A architecture.

%%%%%%%%%%%%%%%%%%%%%%%%%%%%%%%%%%%%%%%%%%%%%%%%%%%%%%%%%%%%%%%%%%%%%%%%%%%%%%%%%%%%%%
\section{Data-Driven Modeling and Training}
In this section, we detail the process of constructing the training dataset from parametric partial differential equation (PDE) solutions and present the methodology employed to train the proposed MI2A architecture. 
We begin by establishing the notation used to represent our data and describe the procedure for generating overlapping input–target pairs for sequence-to-sequence prediction tasks. This is followed by a description of the preprocessing steps, including data normalization and batch partitioning. We then provide a detailed overview of our MI2A architecture, comprising a denoising convolutional autoencoder for latent representation learning and the MI2A evolver for modeling the temporal dynamics in the latent space. Finally, we introduce a mean squared error decomposition, which distinguishes between dissipation (amplitude) and dispersion (phase) errors in wave propagation phenomena.

\subsection{Snapshot Representation and Data Structure}
Consider a dataset consisting of solution snapshots generated from partial differential equation (PDE) simulations conducted over a range of physical parameters. Each snapshot represents the system state at a given time and parameter configuration, forming a high-dimensional, temporally and parametrically varying dataset suitable for training data-driven models of dynamical systems:
\[
\mathcal{D} = \bigl\{\mathbb{U}_{\mu_1}, \ldots, \mathbb{U}_{\mu_{N_\mu}}\bigr\} \;\in\; \mathbb{R}^{N_\mu \times N_T \times N},
\]
where \(N_\mu\) denotes the number of sampled physical parameters, \(N_T\) is the number of time steps, and \(N\) is the spatial dimension. For each parameter \(\mu_i\), we store the snapshots of the solution in
\[
\mathbb{U}_{\mu_i} \;=\; \{\mathbf{U}_{N,\mu_i}^{(1)}, \ldots, \mathbf{U}_{N,\mu_i}^{(N_T)}\}
\;\in\; \mathbb{R}^{N_T \times N},
\]
so that \(\mathbf{U}_{N,\mu_i}^{(t)}\) represents the PDE solution in time \(t\) over an \(N\)-dimensional spatial mesh.

\subsection{Preprocessing and Normalization}
To facilitate sequence-to-sequence prediction, each trajectory $\mathbb{U}_{\mu_i}$ is partitioned into overlapping input-target pairs. Let $N_t$ denote the length (in time steps) of the model's input sequence. We define:
$$
N_s=N_T-2 N_t+1
$$
as the number of overlapping segments of length $2 N_t$ that can be extracted from a single trajectory. Within each segment, the first $N_t$ time steps are used as the input sequence $\mathbf{X}_{\mathrm {Train }}$, and the subsequent $N_t$ time steps constitute the corresponding target sequence $\mathbf{Y}_{\mathrm {Train }}$. This procedure yields $N_s$ input-target pairs for each parameter instance $\mu_i$, resulting in the full training tensors
$$
\mathbf{X}_{\mathrm {Train }}, \mathbf{Y}_{\mathrm {Train }} \in \mathbb{R}^{N_\mu \times N_s \times N_t \times N}
$$
where $N_\mu$ is the number of parameter samples and $N$ denotes the spatial degrees of freedom.
For practical training, the dataset is reshaped into a flat batch format:
$$
\mathbf{X}_{\mathrm {Train }}, \mathbf{Y}_{\mathrm {Train }} \in \mathbb{R}^{N_m \times N_t \times N}, \quad \mathrm { where } \quad N_m=N_\mu \times N_s
$$
so that each training sample corresponds to a pair of temporally aligned input and target sequences.
Before training, each snapshot is scaled to the range \([0,1]\) via min-max normalization:
\[
\overline{X}_{\mathrm{Train}}
= \frac{{X}_{\mathrm{Train}} - {X}_{\mathrm{Train}_{,\min}}}
       {{X}_{\mathrm{Train}_{,\max}}-{X}_{\mathrm{Train}_{,\min}}}
\quad\mathrm{and}\quad
\overline{Y}_{\mathrm{Train}}
= \frac{{Y}_{\mathrm{Train}} - {Y}_{\mathrm{Train}_{,\min}}}
       {{Y}_{\mathrm{Train}_{,\max}}-{Y}_{\mathrm{Train}_{,\min}}},
\]
such that \(\overline{X}_{\mathrm{Train}}, \overline{Y}_{\mathrm{Train}} \in [0,1]^{N_m \times N_t \times N}\). 
This step mitigates the features that dominate the training due to different scales.
In our experiments, we compute a global min and max over all training data $\mathcal{D}$ for consistency between different parameters $\mu_i$.

\subsection{MI2A Architecture and Loss Function}
With the training dataset prepared, we now introduce our denoising convolutional autoencoder and MI2A evolver. The convolutional autoencoder acts as a feature extractor, learning robust and noise-resilient latent representations from snapshot data. These latent features serve as input to the MI2A evolver, which is responsible for modeling their temporal evolution. The architectural details and the temporal integration strategy of the MI2A evolver are presented in Section~\ref{sec:method}.
\subsubsection{Denoising Convolutional Autoencoder Loss}
To promote robust feature extraction, Gaussian noise is added to the normalized training data during the autoencoder training phase. This denoising strategy encourages \cite{vincent2008extracting, deo2022predicting} the encoder to learn invariant and generalizable latent representations that are robust to perturbations in the input
\[
\tilde{X}_{\mathrm{Train}} 
\;=\; 
\overline{X}_{\mathrm{Train}} 
\;+\;
\mathcal{N}(\mathrm{mean}, \mathrm{SD}),
\]
and pass them through a convolutional encoder $f_\theta(\cdot)$, which maps it to a latent representation $\mathbf{Z}$. A decoder $g_\phi(\cdot)$ reconstructs the input, yielding $\hat{X}$. The reconstruction loss is:
\begin{equation}
\mathcal{L}_{\mathrm{AE}}=\left\|\hat{X}-\bar{X}_{\mathrm{Train}}\right\|^2
\end{equation}
which encourages noise removal and feature extraction.
\subsubsection{MI2A Evolver Loss} 
The MI2A evolver advances the learned latent representation in time and reconstructs the predicted solution $X'$ in the physical domain via the decoder. A supervised loss function is then employed to quantify the discrepancy between the predicted output and the ground truth 
$\bar{Y}_{Train}$: 
\begin{equation} 
\mathcal{L}_{\mathrm{evolver}} = \left\| X^{\prime} - \bar{Y}_{\mathrm{Train}} \right\|^2. \end{equation}

To jointly optimize both the autoencoder and the temporal evolver, we define a total loss function that combines the autoencoder reconstruction loss $\mathcal{L}_{\mathrm{AE}}$ and the evolver prediction loss $\mathcal{L}_{\mathrm{evolver}}$: 
\begin{equation} 
\mathcal{L}_{\mathrm{total}} = (1 - \xi)\mathcal{L}_{\mathrm{AE}} + \xi\mathcal{L}_{\mathrm{evolver}}, 
\label{eq:l_total}
\end{equation} 
where $\xi \in [0,1]$ is a tunable hyperparameter that balances the trade-off between latent feature reconstruction and temporal sequence prediction. This combined loss function has been used in our previous works on spatio-temporal modeling \cite{deo2022predicting, gupta2022hybrid, bukka2021assessment}.

\subsection{Dispersion--Dissipation Decomposition of MSE}
In wave prediction tasks, the precise modeling of the amplitude and phase components is crucial to ensure accurate predictions. 
While the MI2A evolver loss effectively minimizes the error between predictions and ground truth, the standard mean squared error does not distinguish between errors arising from amplitude mismatches and those due to phase misalignment. 
This limitation is particularly significant in the context of wave phenomena, where phase errors can lead to incorrect wavefront propagation even if the amplitude is correctly predicted.
To address this issue, we introduce a novel loss decomposition that explicitly separates the total error into amplitude and phase components. Specifically, the total MSE at each time step ($\tau(t_j)$) is decomposed as follows:
\begin{equation}
\begin{aligned}
 \tau(t_j) & = \frac{1}{N}\sum_{i = 0}^{N}\left(\overline{Y}_{\mathrm{Train},i}^{j} - {X}^{'j}_i \right)^2, \quad \forall i \in \Omega,\\
& = \left[\sigma\left(\overline{Y}_{\mathrm{Train}}^{j}\right)-\sigma\left({X}^{'j}\right)\right]^2
+\left(<{\overline{Y}_{\mathrm{Train}}^{j}}>-<{X}^{'j}>\right)^2
+ 2(1-\rho) \sigma\left(\overline{Y}_{\mathrm{Train}}^{j}\right) 
\sigma\left({X}^{'j}\right),
\label{eqn:mse-BV-improved}
\end{aligned}
\end{equation}
where $\overline{Y}_{\mathrm{Train}}^{j}$ is the ground-truth solution and ${X}^{'j}$ is the predicted solution at time-step $t_j$. 
The notation $<\cdot>$ and $\sigma(\cdot)$ denote the spatial means and standard deviations, and $\rho$ represents the correlation coefficient. The complete derivation of this decomposition is given in \ref{appendix:dispersion-dissipation}. 
From this decomposition, we extract two distinct error components:
$$
\begin{aligned}
& \tau_{\mathrm{DISS}}=\left[\sigma\left(\bar{Y}_{\mathrm{Train}}^{j}\right)-\sigma\left(X^{\prime j}\right)\right]^2+\left(<\bar{Y}_{\mathrm {Train }}^{j}>-<X^{\prime j}>\right)^2,
\\
& \tau_{\mathrm{DISP}}=2(1-\rho) \sigma\left(\bar{Y}_{\mathrm{Train}}^{j}\right) \sigma\left(X^{\prime j}\right),
\end{aligned}
$$
where, $\tau_{\mathrm {DISS }}$ measures dissipation (amplitude loss), and $\tau_{\mathrm {DISP }}$ quantifies dispersion (phase mismatch). In particular, when $\rho=1$, the phase error is zero, resulting in $\tau_{\mathrm {DISP }}=0$ . 

To explicitly account for both amplitude and phase error in the training loss, we incorporate this decomposition into the evolver loss:
\begin{equation}
\mathcal{L}_{\mathrm{evolver}} =  
 \psi\,\tau_{\mathrm{DISP}} + (1-\psi)\,\tau_{\mathrm{DISS}},
\label{eqn:loss_function_full}
\end{equation}
where $\psi$ adjusts the trade-off between phase correction (dispersion) and amplitude correction (dissipation). Empirically, we find that prioritizing phase correction (i.e., choosing a larger $\psi$) improves long-term stability in wave predictions. The final training objective is obtained by adding Eq. \ref{eqn:loss_function_full} in the total loss function $\mathcal{L}_{\mathrm{total}}$ Eq.~\ref{eq:l_total}.

\subsection{Forward Pass and Implementation}
During each training iteration, the model performs two sequential forward passes: the first through the denoising convolutional autoencoder to obtain latent representations, and the second through the MI2A evolver to model their temporal evolution. A combined loss, incorporating both reconstruction and predictive components, is then computed and used to update the model parameters via backpropagation. The overall training procedure is summarized in Algorithm~\ref{alg:alg1}.

We first sample a mini-batch $\tilde{X}_{\mathrm{Train}}^{b} \subset \tilde{X}_{\mathrm{Train}}$ (noisy data). The encoder maps this batch to latent representations $\mathbf{Z}^b$, which the decoder reconstructs into $\hat{X}^b$. A denoising autoencoder loss $\mathcal{L}_{\mathrm{AE}}$ then measures reconstruction error against the uncorrupted $\overline{X}_{\mathrm{Train}}^{\,b}$.  
Next, we pass these latent states $\mathbf{Z}^b$ through the MI2A evolver, yielding an evolved latent state $\mathbf{Z}'^b$. We decode $\mathbf{Z}'^b$ back to the physical domain, obtaining ${X}'^{b}$. Comparing ${X}'^{b}$ to $\overline{Y}_{\mathrm{Train}}^{\,b}$ yields the evolver loss $\mathcal{L}_{\mathrm{evolver}}$. We form the total loss $\mathcal{L}_{\mathrm{total}}$ by weighting both losses (using Eq. ~\ref{eq:l_total}), and backpropagate to jointly update all parameters (encoder, decoder, evolver). 
\begin{algorithm}[H]
\caption{\textbf{MI2A Training Algorithm}}
\label{alg:alg1}
\begin{algorithmic}[1]
\STATE \textbf{Input:} $\tilde{X}_{\mathrm{Train}}, \overline{X}_{\mathrm{Train}}, \overline{Y}_{\mathrm{Train}}, N_{\mathrm{epochs}}, N_{b}, \xi, \psi$
\STATE \textbf{Output:} $\theta^* = \{\theta_{Enc}^*, \phi_{Dec}^*, \theta_{\mathrm{evolver}}^*\}$
\STATE Initialize parameters $\theta = \{\theta_{Enc}, \phi_{Dec}, \theta_{\mathrm{evolver}}\}$
\FOR{epoch $= 1$ to $N_{\mathrm{epochs}}$}
    \STATE \textbf{Sample a batch} $\tilde{X}_{\mathrm{Train}}^{b} \subset \tilde{X}_{\mathrm{Train}}$
    \STATE \textbf{Z}$^{b} \leftarrow f_{\theta}(\tilde{X}_{\mathrm{Train}}^{b}; \theta_{Enc})$ \hfill // Encoder forward pass
    \STATE $\hat{\mathcal{X}}^{b} \leftarrow g_{\phi}(\mathbf{Z}^{b}; \phi_{Dec})$ \hfill // Decoder forward pass
    \STATE \textbf{Z}$^{\prime b} \leftarrow \Phi(\mathbf{Z}^{b}; \theta_{\mathrm{evolver}})$ \hfill // Evolver forward pass
    \STATE $\mathcal{X}^{\prime b} \leftarrow g_{\phi}(\mathbf{Z}^{\prime b}; \theta_{Dec})$ \hfill // Decode evolved latent
    \STATE $\mathcal{L} \leftarrow \mathrm{ComputeLoss}(\hat{\mathcal{X}}^{b}, \overline{X}_{\mathrm{Train}}^{b}, \mathcal{X}^{\prime b}, \overline{Y}_{\mathrm{Train}}^{b}, \xi, \psi)$ \hfill // Use Eq. \ref{eq:l_total}
    \STATE $\hat{\mathbf{g}} \leftarrow \nabla_\theta \mathcal{L}$ \hfill // Backpropagation
    \STATE $\theta \leftarrow \mathrm{ADAM}(\theta, \hat{\mathbf{g}})$ \hfill // Parameter update
\ENDFOR
\RETURN $\theta^* = \{\theta_{Enc}^*, \phi_{Dec}^*, \theta_{\mathrm{evolver}}^*\}$
\end{algorithmic}
\end{algorithm}

Overall, this procedure jointly learns a robust latent representation and a powerful time-evolution operator in the latent space. 
By repeatedly minimizing the weighted combination of reconstruction and prediction losses, the MI2A framework (via encoder, decoder, and evolver) can model the complex dynamics of parametric PDEs while controlling for dispersion and dissipation errors.

\section{Numerical Results}
In this section, we discuss how the proposed architecture can predict the evolutionary behavior of hyperbolic PDEs. 
The effectiveness of the proposed methodology will be demonstrated by solving the 1D linear convection equation, the 1D nonlinear viscous Burgers' equation, and the 2D Saint-Venant shallow water system.
\subsection{Linear convection equation}
We first examine the linear convection equation posed on the one-dimensional domain $\Omega=[0,1]$. Let $U$ denote the solution to the parametric PDE:
\begin{equation}
\frac{\partial U}{\partial t}+\mu \frac{\partial U}{\partial x}=0
\end{equation}
subject to the initial condition
\begin{equation}
U(x, 0)=U_0(x) \equiv f(x)
\end{equation}
where $\mu \in[0.775,1.25]$ is the wave phase speed. We choose
\begin{equation}
f(x)=\frac{1}{\sqrt{2 \pi \rho}} \exp \left(-\frac{x^2}{2 \rho}\right),
\end{equation}
with $\rho=10^{-4}$. Although $\rho$ can be set to any other positive value for generality, this choice of $\rho$ represents a practical small-variance Gaussian.
Since the exact solution is $U(x, t)=f(x-\mu t)$, we use the closed-form expression to generate ground-truth data in the space-time domain $[0,1] \times[0,1]$. We discretize the spatial domain into 256 grid points and sample 200 equally spaced time steps, resulting in snapshots of dimension $256 \times 200$. We consider $N_\mu=20$ parameter instances $\left\{\mu_{\mathrm{train}, i}\right\}$ uniformly distributed over $[0.775,1.25]$ for training and $N_{\text {test }}=19$ parameter instances for testing defined by $\mu_{\text {test }, i}=$ $\frac{1}{2}\left(\mu_{\text {train }, i}+\mu_{\text {train }, i+1}\right)$.

For this test case, we employ a 17-layers MI2A network. The encoder consists of convolutional layers, max-pooling operations, and fully connected layers, ultimately reducing the snapshot dimension to a latent space of size $r$. The encoder, evolver, and decoder architectural specifics are summarized in Tables~\ref{tab:LC_ENC}, \ref{tab:evolver} and \ref{tab:LC_Dec} .
The total number of trainable parameters is $203,557$. Training was performed from scratch in TensorFlow on a NVIDIA RTX 6000 Ada GPU, converging after 1500 epochs in approximately 20 minutes of wall-clock time.

\begin{figure*}
\centering
\includegraphics[width=\textwidth]{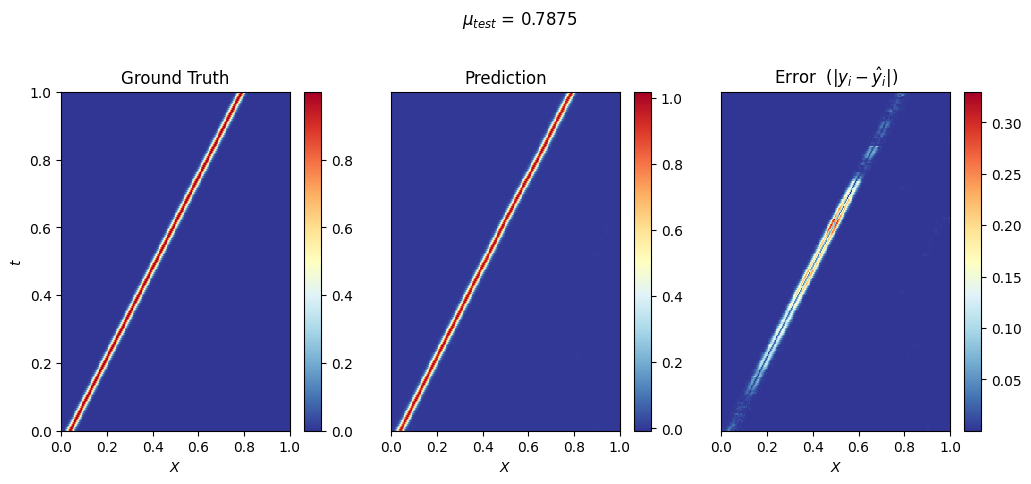}
 \caption{Linear convection problem: Exact solution (left), MI2A solution with n = 2 (center) and error $ e = |\hat{u} - u|$ (right) for the testing parameter $\mu_{test}$ = 0.7875 in the space-time domain.}
\label{fig:char_ABCRAN_mu_0}
\end{figure*}

In this study, the first ten time steps of data from the linear convection equation with $\mu = 0.7875$ are provided as input to the MI2A architecture. The nonlinear reduced manifold dimension is set to $r=2$ for this case, with $\psi$ fixed at 0.7 during training (parameter weighing dissipation and dispersion components in $\mathcal{L}_{\mathrm{evolver}}$). Figure~\ref{fig:char_ABCRAN_mu_0} presents a comparison between the exact solution and the MI2A approximation for this specific test parameter. The MI2A model with $r=2$ effectively captures the amplitude and accurately predicts the wave velocity.

\begin{table*}[htbp]
\centering
\caption{Detailed attributes of convolutional and dense layers in the encoder \( f_{\theta}(\cdot;\theta) \).}
\label{tab:LC_ENC}
\resizebox{\textwidth}{!}{%
\begin{tabular}{clllcccc}
\toprule
\textbf{Layer} & \textbf{Layer Type} & \textbf{Input Dimension} & \textbf{Output Dimension} &\textbf{ Kernel Size} &\textbf{ \# Filters/Neurons }& \textbf{Stride} \\
\midrule
1 & Conv 1D   & $(N_T,256,1)$ & $(N_T,128,64)$ & 5 & 64 & 2 \\
  & MaxPool 1D & $(N_T,128,64)$ & $(N_T,64,64)$ & -- & -- & -- \\
2 & Conv 1D   & $(N_T,64,64)$ & $(N_T,32,32)$ & 5 & 32 & 2 \\
  & MaxPool 1D & $(N_T,32,32)$ & $(N_T,16,32)$ & -- & -- & -- \\
  & Flatten    & $(N_T,16,32)$ & $(N_T,512)$   & -- & -- & -- \\
3 & Dense      & $(N_T,512)$   & $(N_T,128)$   & -- & 128 & -- \\
4 & Dense      & $(N_T,128)$   & $(N_T,64)$    & -- & 64 & -- \\
5 & Dense      & $(N_T,64)$    & $(N_T,r)$     & -- & $r$ & -- \\
\bottomrule
\end{tabular}
}
\end{table*}

\begin{table*}[htbp]
\centering
\renewcommand{\arraystretch}{1.2}
\setlength{\tabcolsep}{8pt}
\caption{Attributes of the MI2A Evolver Function $\Phi(.;\theta_{evolver})$}
\label{tab:evolver}
\resizebox{\textwidth}{!}{%
\begin{tabular}{@{} p{0.9cm} p{3.5cm} p{3.0cm} p{3.0cm} p{5.4cm} @{}}
\toprule
 \textbf{Layer} & \textbf{Layer Type} & \textbf{Input Dimension} & \textbf{Output Dimension} & \textbf{Comments} \\
\midrule
6 & RNN-LSTM-Encoder-1 & (None, $N_T$, r) & (None, $N_T$, p) & LSTM with p units \\
7 & RNN-LSTM-Encoder-2 & (None, $N_T$, p) & (None, $N_T$, p) & Two layers LSTM \\
& RNN-Decoder Input & (None, p) expanded to (None, $N_T$, p) & (None, $N_T$, p) & Repeat vector of encoder output for decoding, Using Eq.~\ref{eqn:dec_inp} \\
8 & RNN-LSTM-Decoder-1 & (None, $N_T$, p) & (None, $N_T$, p) & Decoding with p units \\
9 & RNN-LSTM-Decoder-2 & (None, $N_T$, p) & (None, $N_T$, p) & Second LSTM decoder \\
10 & Learnable Attention Weights & (None, $N_T$, p) & (None, $N_T$, p) & Learnable weights via dense layer \\
& Attention Dot-product & [(None,$N_T$,p), (None,$N_T$,p)] & (None, $N_T$, $N_T$) & Dot product for attention map \\
& Softmax Activation & (None, $N_T$, $N_T$) & (None, $N_T$, $N_T$) & Attention map with weights summing to 1 \\
& Context Vector & [(None,$N_T$,$N_T$), (None,$N_T$,p)] & (None, $N_T$, p) & Context computed from attention mechanism \\
11 & Input Sequence Derivative & (None, $N_T$, p) & (None, $N_T$, p) & Estimated derivative via Conv1D  \\
& Decoder Skip-connection & [(None,$N_T$,p)]$\times$3 & (None, $N_T$, p) & Combines context, derivative, and decoder hidden states \\
12 & RNN-Decoder Output & (None, $N_T$, p) & (None, $N_T$, r) & Final dense layer projects to latent dimension \\
\bottomrule
\end{tabular}
}
\end{table*}

\begin{table*}[htbp]
\centering
\caption{Attributes of transpose convolutional and dense layers in the decoder $g_\phi(.;\phi)$.}
\label{tab:LC_Dec}
\resizebox{\textwidth}{!}{%
\begin{tabular}{clllcccc}
\toprule
\textbf{Layer} & \textbf{Layer Type} & \textbf{Input Dimension} & \textbf{Output Dimension} &\textbf{ Kernel Size} &\textbf{ \# Filters/Neurons }& \textbf{Stride} \\
\midrule
     13 & \text{Dense} & {($N_T$, r)} & {($N_T$, 64)} & - & 64 & - \\
    14 & \text{Dense} &{($N_T$, 64)} & {($N_T$, 128)} & - & 128 & - \\
    15 & \text{Dense} &{($N_T$, 128)} & {($N_T$, 512)} & -  & 512 & -\\
     & \text{Reshape} &{($N_T$, 512)} & {($N_T$, 16, 32)} & -  & - & -\\
     & \text{UpSampling 1D} &{($N_T$, 16, 32)} & {($N_T$, 32, 32)} & -  & - & -\\
    16 & \text{Conv 1D Transpose} & {($N_T$, 32, 32)} & {($N_T$, 64, 64)} & [5]& 64&2 \\
     & \text{UpSampling 1D} &{($N_T$, 64, 64)} & {($N_T$, 128, 64)} & -  & - & -\\
    17 & \text{Conv 1D Transpose} & {($N_T$, 128, 64)} & {($N_T$, 256, 1)} & [5]& 1&2 \\
\bottomrule

    \end{tabular}  
    }
    \end{table*}

\subsubsection{Impact of MI2A on time series prediction}
\begin{figure*}
\centering
\includegraphics[width=\textwidth]{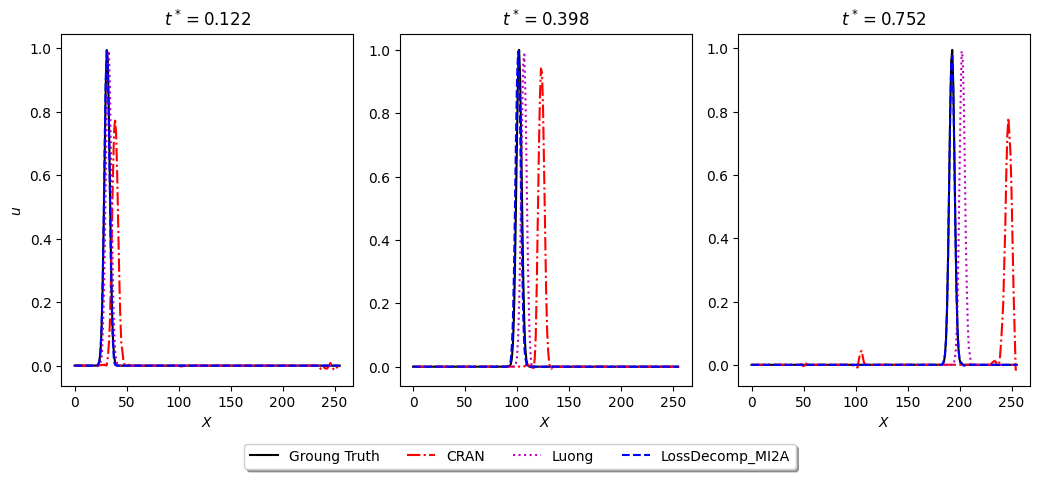}
 \caption{Linear convection problem: Comparison of full-order model solution, MI2A, CRAN, and Luong-based ROM solution at three time instants ($t*=[0.122,0.398, 0.752]$), where non-dimensional time $t^*= t \mu/L$. 
 %Here n = 2 for test prediction for Case 1 with parameters...
}
\label{fig:FOM_ABCRAN_CRAN_MU0}
\end{figure*}

To evaluate the effectiveness of the MI2A architecture, we compare its predictive performance against two alternative reduced-order models that employ different temporal evolution strategies. The first model, CRAN, consists of a convolutional autoencoder for dimensionality reduction paired with a sequence-to-sequence RNN-LSTM evolver to capture temporal dynamics. The second model also utilizes a convolutional autoencoder but replaces the RNN-LSTM with Luong’s attention mechanism, which adaptively weights encoder states to enhance long-range dependencies. 
This comparison allows us to assess whether the MI2A framework offers advantages in predictive accuracy and stability over existing reduced-order modeling approaches.
For this comparison, we select a wave phase speed of 0.7875 and evaluate the capability of each method to accurately predict wave propagation. 
The prediction accuracy is quantified using three error metrics: mean squared error (MSE), mean absolute error (MAE), and maximum error ($L_\infty$), defined as follows:
\begin{equation}
\operatorname{MSE}(\mathbf{u}, \hat{\mathbf{u}})=\sum_{i=1}^{N}\frac{(\hat{\mathbf{u}}_i^j-\mathbf{u}_i^j)^{2}}{N},
\label{eqn:mse}
\end{equation}
\begin{equation}
\operatorname{MAE}(\mathbf{u}, \hat{\mathbf{u}})=\sum_{i=1}^{N}\frac{|\hat{\mathbf{u}}_i^j-\mathbf{u}_i^j|}{N},
\label{eqn:mae}
\end{equation}
\begin{equation}
\operatorname{L_\infty}(\mathbf{u}, \hat{\mathbf{u}})=\operatorname{max}(|\hat{\mathbf{u}}_i^j-\mathbf{u}_i^j|),
\label{eqn:max}
\end{equation}
here $N$ is the spatial degrees of freedom.
Figure~\ref{fig:FOM_ABCRAN_CRAN_MU0} compares model predictions at selected nondimensional time instances, defined as $t^*=\frac{t \mu}{L}$, specifically at $t^*=0.122,0.398$, and 0.752 . We employ a sequence-to-sequence framework to forecast sequences of ten time steps, treating each as an individual time horizon. As a result, the third, tenth, and nineteenth time horizons correspond to predictions at the thirtieth, hundredth, and one hundred ninetieth time steps, respectively. The results in Fig.~\ref{fig:FOM_ABCRAN_CRAN_MU0} demonstrate that the MI2A model consistently predicts wave velocity and peak amplitude with high accuracy across all evaluated test steps.
Conversely, the CRAN architecture utilizing a standard LSTM sequence-to-sequence evolver exhibits difficulties in capturing wave propagation accurately beyond the initial time horizon. Similarly, the reduced-order model employing Luong's attention mechanism struggles to precisely predict the wave propagation phase.

\begin{figure*}
\centering
\includegraphics[width=\textwidth]{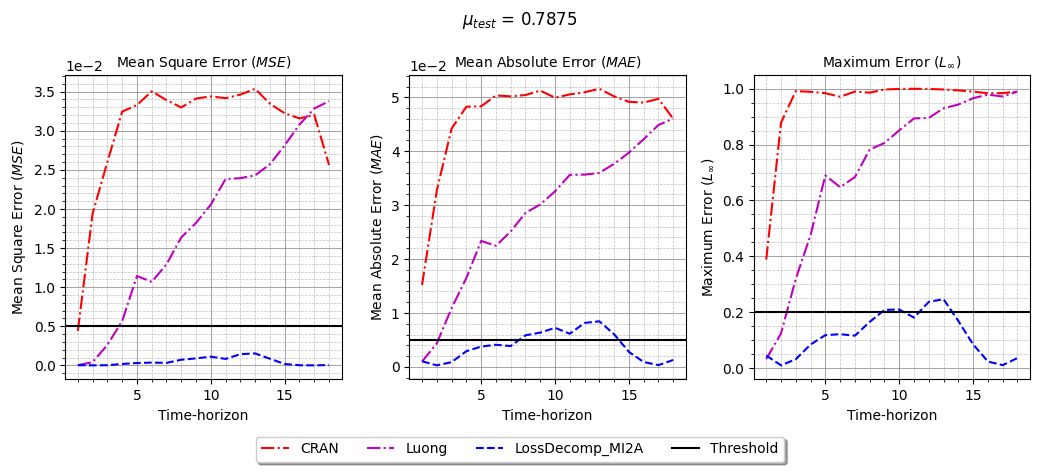}
 \caption{Linear convection problem: Comparison of mean squared error, mean absolute error, and maximum error of MI2A with CRAN and Luong network
}
\label{fig:error_MU0}
\end{figure*}

Figure~\ref{fig:error_MU0} presents a comparative analysis of the mean squared error (MSE), mean absolute error (MAE), and $L_\infty$ error norms for the CRAN, Luong-based ROM, and MI2A models. The MI2A model exhibits consistently lower MSE, MAE, and $L_\infty$ errors compared to both the CRAN and Luong-based approaches. These results highlight the superior accuracy of the MI2A network in reducing errors for the linear convection equation.
Notably, the MSE in MI2A predictions remains negligible throughout the entire prediction period, in stark contrast to the CRAN and Luong models. Furthermore, both the MSE and maximum error ($L_\infty$) in the CRAN and Luong models increase over time, indicating error accumulation as the prediction horizon extends. In contrast, the MI2A model maintains an error below the threshold, demonstrating its robustness and stability. This confirms that MI2A effectively learns and models the linear convection equation with greater accuracy than the CRAN and Luong networks.

%%%%%%%%%%%%%%%%%%%%%%%%%%%%%%%%%%%%%%%%%%%%%%%%%%%%%%%%%%%%%%%%%%%%%%
\subsubsection{Effect of loss decomposition}
\begin{figure*}
\centering
\includegraphics[width=\textwidth]{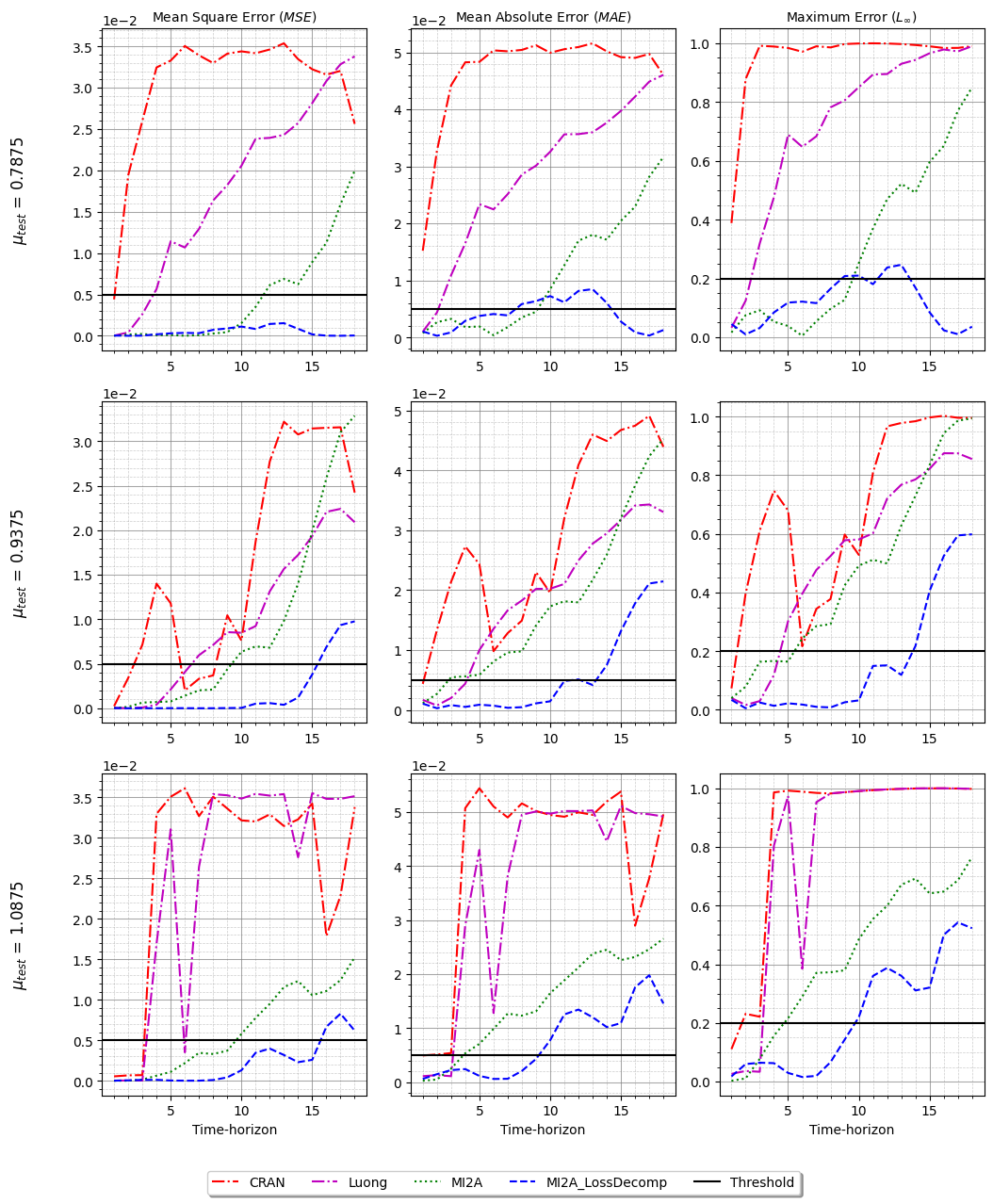}
 \caption{Linear convection problem: Comparison of generalisation error across the parameter space $\mu \in [0.7875, 1.2375]$. }
\label{fig:LC_denoising_generalisation_error}
\end{figure*}

In this section, we investigate the impact of decomposing the mean squared error loss into dissipation and dispersion components during MI2A training and assess its influence on generalization in varying parameter regimes. 
A key benefit of loss decomposition is the intrinsic regularization effect, which mitigates overfitting and enhances predictive accuracy in unseen parameter instances. 
To demonstrate this advantage, predictions were evaluated for three distinct test cases using the network trained with loss decomposition and with standard mean-square error evolver training loss ($\mathcal{L}_{\mathrm{evolver}}$). Specifically, Test Case 1 corresponds to $\mu = 0.7875$, showing a wave phase speed less than unity; Test Case 2 corresponds to $\mu = 0.9375$, with a wave phase speed approaching unity; and Test Case 3 corresponds to $\mu = 1.0875$, representing a wave phase speed greater than unity. 
Figure (\ref{fig:LC_denoising_generalisation_error}) illustrates that the MSE, MAE, and $L_\infty$ error norms for all three test cases are consistently lower for MI2A trained with loss decomposition compared to MI2A, CRAN and Luong networks with mean squared error loss in evolver, clearly indicating superior generalization capability. Thus, the proposed loss decomposition improves model performance across the entire parameter space and extends the predictive horizon.

%, mean absolute error (Eq. (\ref{eqn:mae})) a
In addition to the mean squared error (Eq. (\ref{eqn:mse})), and the maximum error (Eq. (\ref{eqn:max})), we consider the time average value of the mean squared error and the maximum errors as an alternative metric to assess the accuracy of the predictions for different parameters, which are given by
\begin{equation}
 <MSE(\mathbf{u},\hat{\mathbf{u}})>=\sum_{j=1}^{N_t}\left(\sum_{i=1}^{N}\frac{(\hat{\mathbf{u}}_i^j-\mathbf{u}_i^j)^{2}}{N}\right)/({N_t}),
\label{eqn:TVmse}
\end{equation}
and
\begin{equation}
<{L_\infty}(\mathbf{u}, \hat{\mathbf{u}})>=\sum_{j=1}^{N_t}\frac{\operatorname{max}(|\hat{\mathbf{u}}_i^j-\mathbf{u}_i^j|)}{N_t},
\label{eqn:TVmax}
\end{equation}
where $N$ is the spatial degrees of freedom, and $N_t$ is number of time steps, and $< >$ denotes the time averaging.
In summary, it can be seen in Table \ref{table:mse_Linf_LC} that the MI2A reduces the mean squared error by an order of magnitude compared to the CRAN and Luong network while it reduces the maximum error by three times.
% {As a function of the data size, a comparison of time averaged mean squared error for training, validation, and test data-set is summarized in Table \ref{table:amt_data_LC}.}

\begin{table}[h]
    \centering
    \caption{Comparison of time-averaged mean squared error (\(\langle MSE \rangle\)) and maximum error (\(\langle L_{\infty} \rangle\)) for the linear convection problem computed using different methods. \colorbox{orange}{ } indicates the best performance.}
    \label{table:mse_Linf_LC}
    \resizebox{\textwidth}{!}{%
    \begin{tabular}{c |c|c|c|c| c|c|c|c}
        \Xhline{3\arrayrulewidth}
        Parameter  & \multicolumn{4}{c|}{$\langle MSE \rangle$} & \multicolumn{4}{c}{$\langle L_{\infty} \rangle$} \\
        % \cmidrule(lr){2-5} \cmidrule(lr){6-9}
        \Xhline{1\arrayrulewidth}
        $\mu$ & MI2A\_LossDecomp & MI2A & Luong & CRAN & MI2A\_LossDecomp & MI2A & Luong & CRAN \\
        \Xhline{2\arrayrulewidth}
        0.7875  & \colorbox{orange}{0.000512} & 0.005142 & 0.018268 & 0.030942 & \colorbox{orange}{0.117071} & 0.332826 & 0.730119 & 0.953459 \\
        % \hline
        0.9375  & \colorbox{orange}{0.002007} & 0.010062 & 0.010310 & 0.016191 & \colorbox{orange}{0.179185} & 0.494597 & 0.544545 & 0.661146 \\
        % \hline
        1.0875  & \colorbox{orange}{0.002092} & 0.006124  & 0.026448 & 0.025267 & \colorbox{orange}{0.224535} & 0.414124 & 0.793614 & 0.851936 \\
        \Xhline{3\arrayrulewidth}
    \end{tabular}
    }
\end{table}

\subsection{Viscous Burgers' equation}
In this section, we examine the viscous Burgers' equation as a model for nonlinear wave propagation. The governing equation is given by: 
\begin{equation}
\frac{\partial{u}}{\partial{t}}+u \frac{\partial u}{\partial x}=\nu \frac{\partial^{2} u}{\partial x^{2}},
\end{equation}
where $\nu$ is the viscosity parameter. The system is subject to Dirichlet boundary conditions and the following initial condition:
\begin{align}
    u(x, 0)&=\frac{x}{1+\sqrt{\frac{1}{t_{0}}} \exp \left(\operatorname{Re} \frac{x^{2}}{4}\right)} \quad \text{on}\quad [0, L], \\ u(0, t)&=u(L, t)=0.
\end{align}
We define the Reynolds number as $R e=\frac{1}{\nu}$, which varies in the range $R e \in[1000,4000]$. A total of $N_{R e}=7$ training parameter instances are selected uniformly across this range. The spatial domain is set as $L=1$ and discretized into 256 grid points, while the temporal domain extends to $t_{\max }=2$ and is discretized into 200 time steps.
The analytical solution corresponding to the given initial condition is expressed as:
\begin{equation}
    u(x, t)=\frac{\frac{x}{t+1}}{1+\sqrt{\frac{t+1}{t_{0}}} \exp \left(\operatorname{Re} \frac{x^{2}}{4 t+4}\right)},
\end{equation}
where $t_0=\exp (R e / 8)$. Due to the convection-dominated nature of the problem, the viscous Burgers' equation can give rise to sharp gradients and, in the limiting case, discontinuous solutions.

\begin{figure*}
\centering
\includegraphics[width=\textwidth]{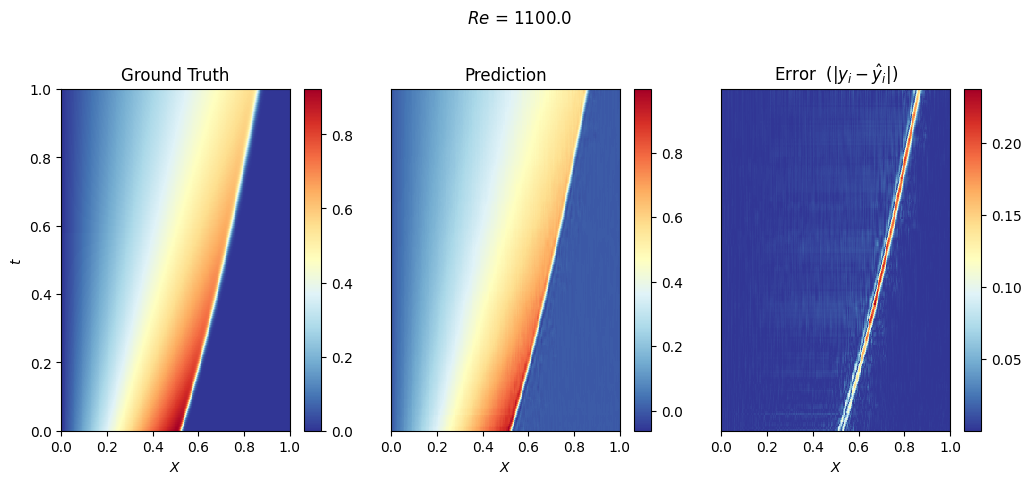}
 \caption{Nonlinear viscous Burgers' problem: Exact solution (left), MI2A solution with n = 2 (center) and error $ e = |y_i - \hat{y}_i|$ (right) for
the testing-parameter instance $Re$ = 1100 in the space-time domain}
\label{fig:Char_ABCRAN_VB_Re_1100}
\end{figure*}

In this test case of nonlinear convection, the neural network architecture with  17-layers of MI2A is used. We also set the dimension of the reduced manifold to two {($r=2$)}. The $\psi$ was set to 0.70.
As input, the first ten time steps of the viscous Burgers' equation with $Re = 1100$ are used.  Figure \ref{fig:Char_ABCRAN_VB_Re_1100} shows both the exact solution and the MI2A approximation for this particular instance of the testing parameter. MI2A solution with {$r=2$} accurately captures the nonlinear wave propagation. 

\subsubsection{MI2A predictions for varying Reynolds number}
\begin{figure*}
  \centering
  % include first image
  
  \includegraphics[width= \linewidth]{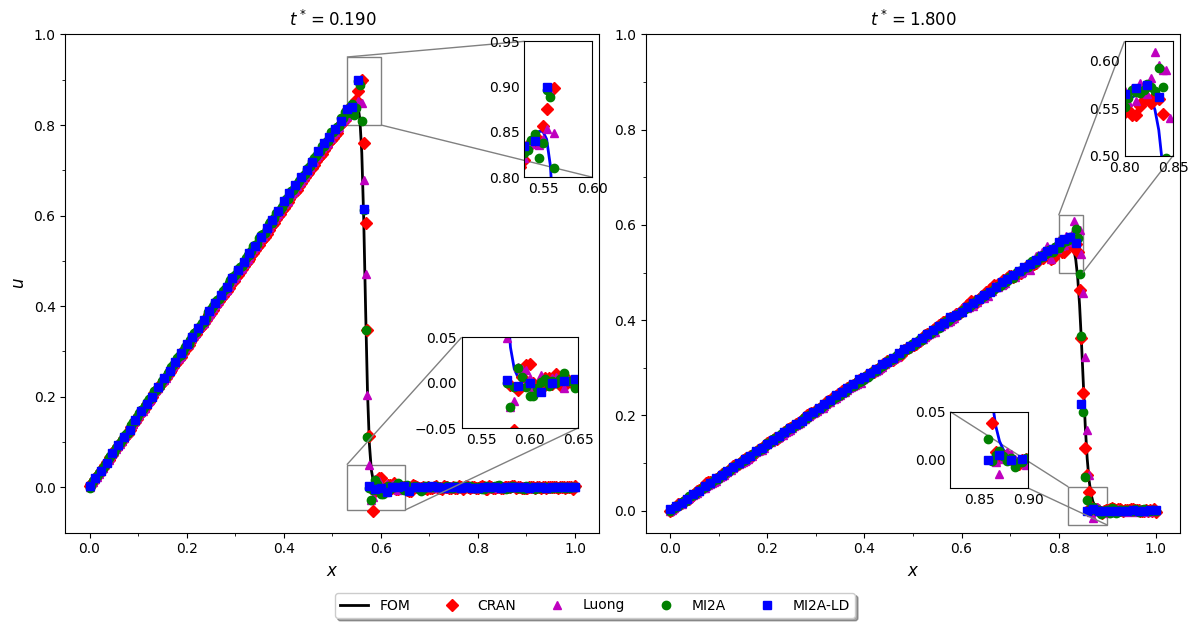}  
\\ (a) \\
  % include third image
  \includegraphics[width = \linewidth]{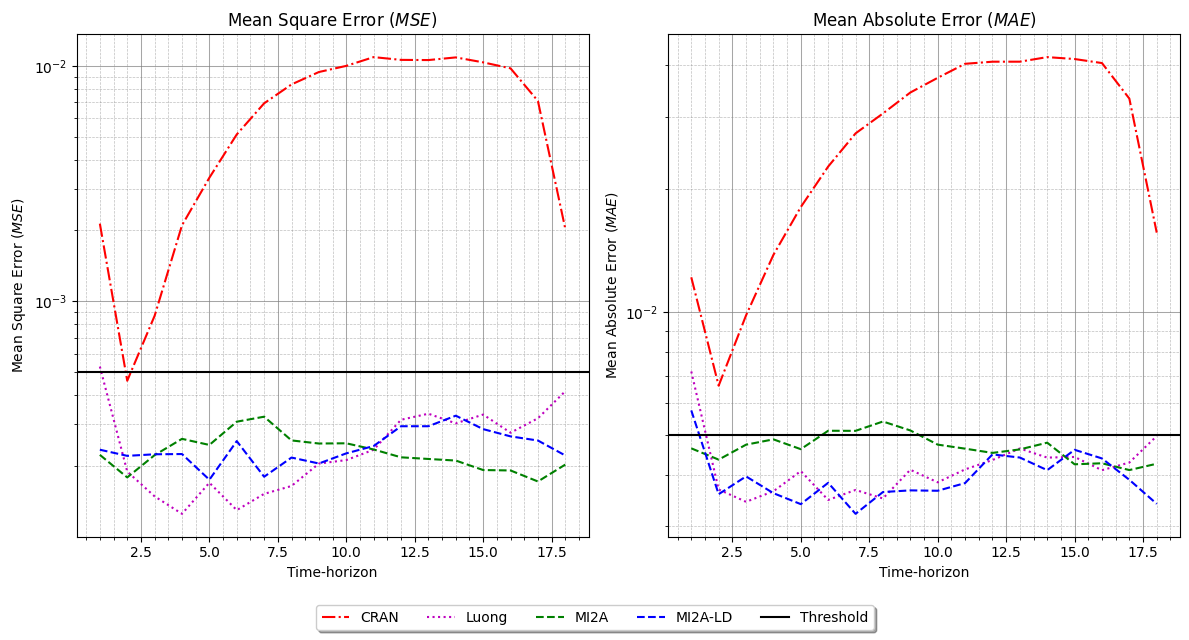}  
\\  (b)
\caption{Error plots and predictions from MI2A and other networks for Re = 1100. (a) Shows models prediction for Re = 1100 at time steps $t^*=\{0.19,1.8\}$. (b) Illustrates mean squared error and mean absolute error from different models for Re = 1100.}
\label{fig:fig_Visc_Burgers_Re_1100}
\end{figure*}

In this section, we evaluate the predictive performance of our MI2A framework in a range of Reynolds numbers (1000 to 4000), which influence the complexity of nonlinear wave dynamics. To assess its robustness, we analyze two representative cases: Test Case 1 (Re = 1100) and Test Case 2 (Re = 3600).
Figure~\ref{fig:fig_Visc_Burgers_Re_1100} compares the predicted values and corresponding errors for Test Case 1 ($Re=1100$). The results indicate that MI2A effectively captures both nonlinear wave propagation and discontinuous features of the solution. In contrast, both the CRAN and Luong architectures exhibit oscillatory behavior near the discontinuity, highlighting MI2A’s superior ability to capture the underlying physics of the viscous Burgers’ equation.

\begin{figure*}
  \centering

  \includegraphics[width=\linewidth]{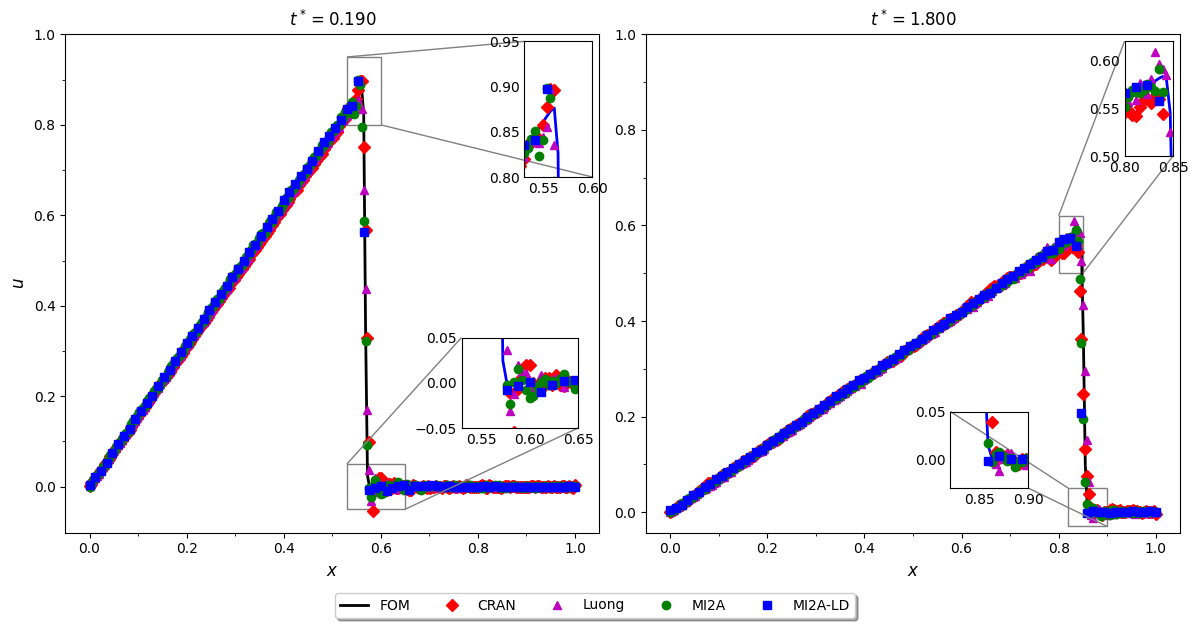}  
\\ (a) \\
  % include third image
  \includegraphics[width=\linewidth]{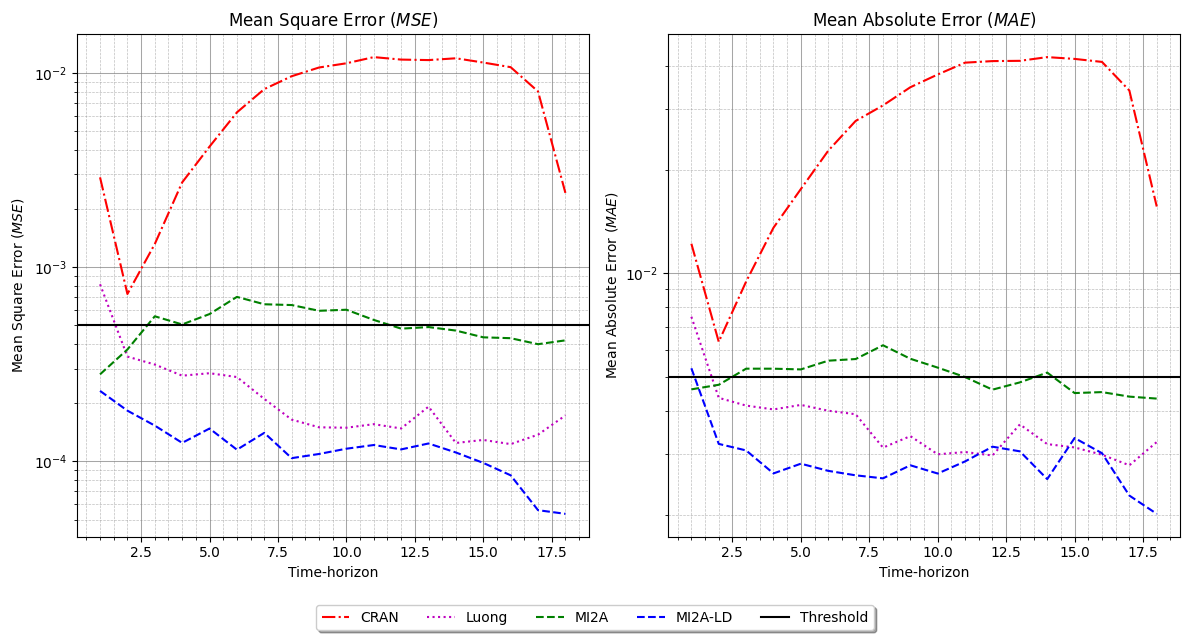}  
\\ (b)

\caption{Nonlinear viscous Burgers' problem: Error plots and predictions from MI2A and other architectures at Re = 3600. (a) Shows different models prediction for Re = 3600 at time steps $t^*=\{0.19,1.8\}$. (b) Illustrates MSE and maximum error from models for Re = 3600.}
\label{fig:fig_Visc_Burgers_Re_3600}
\end{figure*}

To quantify predictive accuracy, we compare the mean squared error (MSE) and mean absolute error (MAE) of MI2A, Luong, and CRAN models, as shown in Fig.~\ref{fig:fig_Visc_Burgers_Re_1100}. MI2A consistently achieves lower errors than CRAN and Luong, reducing prediction errors by approximately $50\%$ for $Re=1100.$
A similar trend is observed in Test Case 2 ($Re=3600$). The MI2A framework accurately models nonlinear wave propagation and discontinuities, whereas CRAN and Luong exhibit oscillatory behavior near sharp gradients. As shown in Fig.~\ref{fig:fig_Visc_Burgers_Re_3600}, MI2A reduces the prediction error of the CRAN and Luong models by approximately $50\%$, demonstrating consistent improvements across different flow regimes.

As summarized in Table \ref{table:mse_Linf_VB}, the MI2A loss decomposition framework achieves the lowest mean squared error (MSE) and mean absolute error (MAE) across all Reynolds numbers, significantly outperforming the CRAN and Luong models. Compared to CRAN, MI2A with loss decomposition reduces MSE by more than an order of magnitude and decreases MAE by a factor of 5 to 10. These substantial improvements highlight MI2A loss decomposition's effectiveness in accurately modeling nonlinear wave propagation while maintaining stability near sharp discontinuities.

\begin{table}[h]
    \centering
    \caption{Comparison of time-averaged mean squared error (\(\langle MSE \rangle\)) and mean absolute error (\(\langle MAE \rangle\)) for the viscous Burgers' problem computed using different methods. \colorbox{orange}{ } indicates the best performance.}
    \label{table:mse_Linf_VB}
    \resizebox{\textwidth}{!}{%
    \begin{tabular}{c | c | c | c | c | c | c | c | c}
        \Xhline{3\arrayrulewidth}
        Parameter  & \multicolumn{4}{c|}{$\langle MSE \rangle$} & \multicolumn{4}{c}{$\langle MAE \rangle$} \\
        \Xhline{1\arrayrulewidth}
        $Re$ & MI2A\_LossDecomp & MI2A & Luong & CRAN & MI2A\_LossDecomp & MI2A & Luong & CRAN \\
        \Xhline{2\arrayrulewidth}
        1100.0  & 0.000233 & \colorbox{orange}{0.000221} & 0.000226 & 0.006584 & \colorbox{orange}{0.003843} & 0.004635 & 0.004382 & {0.028098} \\
        2600.0  & \colorbox{orange}{0.000053} & 0.000330 & 0.000109 & 0.007255 & \colorbox{orange}{0.002317} & 0.004668 & 0.003205 & 0.028112 \\
        4100.0  & \colorbox{orange}{0.000146} & 0.000537 & 0.000239 & 0.007496 & \colorbox{orange}{0.002957} & 0.005060 & 0.003586 & {0.028212} \\
        \Xhline{3\arrayrulewidth}
    \end{tabular}
    }
\end{table}

%%%%%%%%%%%%%%%%%%%%%%%%%%%%%%%%%%%%%%%%%%%%%%%%%%%%%%%%%%%%%%%%%%%%%%%%%%%%%%%%%%%%%%%%%%%%%%%%%%%%%
\subsection{2D Shallow Water Wave Propagation}

In this section, we examine a two-dimensional shallow water model described by the Saint-Venant equations. This system of partial differential equations offers a hydrodynamic framework to compute both the flow velocity and the water level over a two-dimensional domain, incorporating the diverse forces that influence and accelerate the flow. The two-dimensional horizontal Saint-Venant formulation is derived from the vertical integration of the three-dimensional Navier-Stokes equations under the assumptions that the vertical pressure gradient is nearly hydrostatic (an assumption valid for long-wave approximations) and that the horizontal length scale is significantly larger than the vertical length scale.

The Saint-Venant system consists of a mass conservation equation coupled with two momentum conservation equations, and it can be expressed in a non-conservative form as:
\begin{equation}
\left.
\begin{aligned}
\frac{\partial h}{\partial t}+\pder{x}{\left((H+h) u\right)}+\pder{y}{((H+h) v)} &=0, \\
\frac{\partial u}{\partial t}+u \frac{\partial u}{\partial x}+v \frac{\partial u}{\partial y}+g \frac{\partial h}{\partial x}-\nu\left(\frac{\partial^2 u}{{\partial x}^2}+\frac{\partial^2 u}{{\partial y}^2}\right) &=0, \\
\frac{\partial v}{\partial t}+u \frac{\partial v}{\partial x}+v \frac{\partial v}{\partial y}+g \frac{\partial h}{\partial y}-\nu\left(\frac{\partial^2 v}{{\partial x}^2}+\frac{\partial^2 v}{{\partial y}^2}\right) &=0, 
\end{aligned}
\right\}
\end{equation}
where $u$ and $v$ denote the velocities in the $x$ and $y$ directions, respectively, $H$ represents the reference water height, $h$ is the deviation from this reference level, $g$ denotes the gravitational acceleration, and $\nu$ is the kinematic viscosity. Solid wall boundary conditions are imposed along the perimeter of the domain.

\begin{figure*}
\centering
\includegraphics[width=0.8\textwidth]{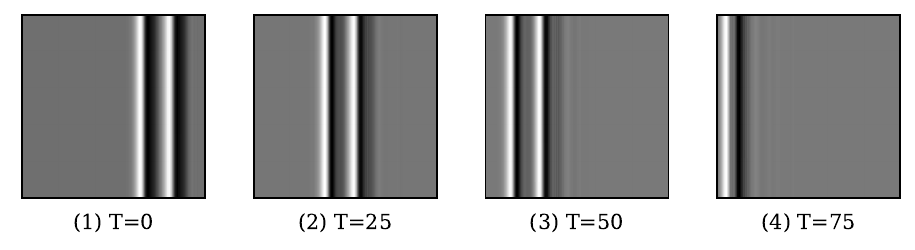}
 \caption{2D Saint-Venant problem: Illustration of the propagation of a plane wave. $T$ represents the number of time steps from the initial condition.}
\label{fig:FOM_PW}
\end{figure*}

A plane wave is employed as the initial condition. The dataset was generated using the Python package TriFlow \cite{nicolas_cellier_2017_584101}, and an example of the evolving wave pattern, as computed by the numerical solver, is presented in Fig. \ref{fig:FOM_PW}. The dataset was assembled by varying the initial position of the plane wave. The temporal domain is set with $t_{\max }=1$ and discretized into 100 time steps, with the resulting images rendered at a resolution of $184 \times 184$ pixels.

\subsubsection{Data-driven predictions via MI2A}

The architecture of the neural network used for this test case is similar to the first two cases. 
In particular, we augment the network architecture to handle the two-dimensional input data and the reduced dimension is set to eight ($r=8$).
 The architecture of the network is identical to the one-dimensional case, only two-dimensional convolution and max-pooling are used instead of one-dimensional operations. 
The total number of trainable parameters (i.e., weights and biases) of the neural network is 4,508,689.
{The model was trained from scratch with TensorFlow \cite{tensorflow2015-whitepaper} using a single NVIDIA RTX 6000 Ada GPU, and 16 cores Intel Xeon w5-3433 CPU with 128GB of system's memory. The training converges in approximately 450 epochs and 3 hours of wall clock time.}

\begin{figure*}
\centering
\includegraphics[width=0.7\textwidth]{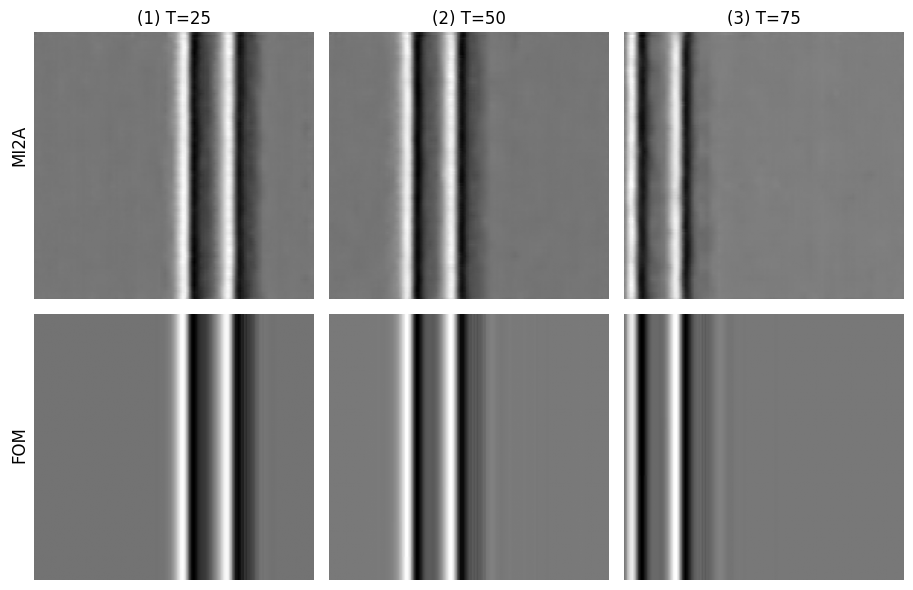}
 \caption{2D Saint-Venant shallow water problem: predicted two-dimensional spatial patterns from MI2A and full-order model solution from a high-fidelity numerical solver.}
\label{fig:pred_2D}
\end{figure*}

We generate a 10 wave solution from the full-order model and used them as input. 
Figure \ref{fig:pred_2D} shows the solution from the full-order model and MI2A architecture. The MI2A framework demonstrates strong predictive capabilities in accurately capturing both the spatial patterns and wave amplitudes of the Saint-Venant equations.
{For a single 2D shallow water simulation, Triflow takes about 5 minutes to generate 100 time steps using a finite difference formulation, while MI2A during inference takes about 15 seconds to generate 100 time-steps providing a speed-up of \textbf{20x}.} 

\begin{figure*}
\centering
\includegraphics[width=0.7\textwidth]{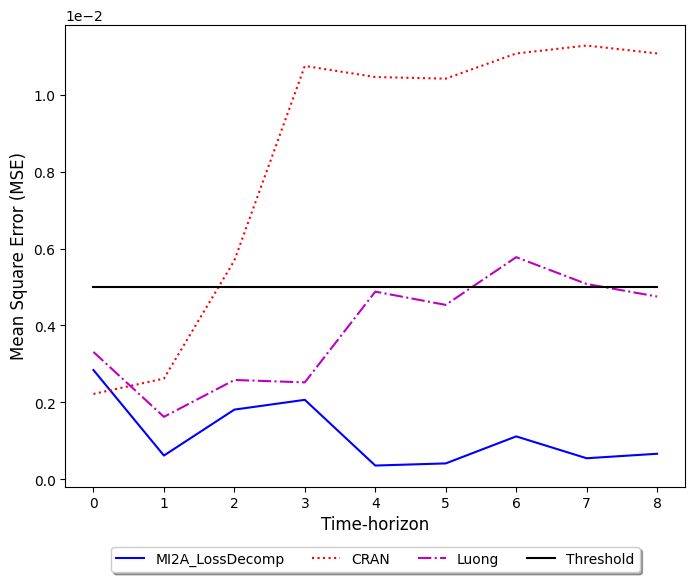}
 \caption{2D Saint-Venant shallow water problem: Comparison of mean square error vs time-horizon for the predicative capability of MI2A, Luong and CRAN. While blue curve is the prediction from MI2A, red curve shows the prediction from CRAN, and mangeta curve shows the prediction from Luong attention ROM. Herein, a sequence of ten time steps is termed as one time-horizon. }
\label{fig:error_2D}
\end{figure*}

In Fig. \ref{fig:error_2D}, the mean squared error of the MI2A, Luong and CRAN predictions {with $r=8$} are compared. 
In comparison to the CRAN and Luong network, the MI2A models exhibit consistently lower mean square error. The results show that the MI2A network substantially outperforms both the CRAN and Luong models by reducing prediction errors in the two-dimensional cases.  
The results suggest that our trained network can perform the wave propagation for the two-dimensional case with minimal hyperparameter tuning, hence that the present algorithm confirms the scalability to multidimensions. 

\section{Conclusion}
In this work, we introduced MI2A (Multi-step Integration Inspired Attention), a novel deep learning framework that integrates multistep numerical integration principles into the attention mechanism for predicting wave propagation governed by hyperbolic PDEs. Unlike conventional deep learning approaches that suffer from accumulated phase and amplitude errors, MI2A dynamically learns multistep weighting coefficients, allowing it to effectively track characteristic trajectories and maintain numerical stability over long time horizons.
A key novel contribution of this study is the loss decomposition in the context of learning-based solvers for wave propagation. Traditional deep learning models typically rely on standard loss functions, such as mean squared error, which do not explicitly separate the fundamental sources of predictive error. In contrast, our proposed framework decomposes the loss into dissipation and dispersion components, thereby enabling the model to independently mitigate phase shifts and amplitude attenuation. This decomposition, inspired by numerical analysis, provides interpretability and control over predictive accuracy, marking a significant advancement in deep learning-based solvers for hyperbolic PDEs.

To evaluate the effectiveness of MI2A, we applied it to three benchmark wave propagation problems of increasing complexity: (i) 1D linear convection eqquation, testing the model's ability to maintain phase accuracy in simple wave dynamics, (ii) 1D nonlinear Burgers equation, demonstrating its capacity to handle nonlinear shock formation, and (iii) 2D Saint-Venant shallow water equations, showcasing its scalability and effectiveness in real-world geophysical flows.
Across all test cases, MI2A outperformed conventional sequence-to-sequence and autoregressive models, successfully mitigating both dissipation and dispersion errors, leading to improved stability and accuracy over extended time horizons. The introduction of loss decomposition as an optimization objective further enhanced MI2A’s capability to learn long-term dependencies while preserving wave properties.
The findings of this work establish MI2A as a general and scalable framework for predicting wave dynamics, with broad applications in computational fluid dynamics, geophysics, climate, and ocean modeling. 
%Moving forward, we will explore applying MI2A to complex multi-scale problems, incorporating adaptive mesh refinement. The proposed approach sets the foundation for future research in interpretable and physics-aware deep learning for scientific computing.

\section*{Acknowledgment}
The present study is supported by Mitacs, Transport Canada, and Clear Seas through the Quiet-Vessel Initiative (QVI) program. 
The authors would like to express their gratitude to Dr. Paul Blomerous and Ms. Tessa Coulthard for their valuable feedback and suggestions.  We also acknowledge that the GPU facilities at the Digital Research Alliance of Canada clusters were used for the training of our deep learning models.
%\end{acknowledgments}
% \printbibliography

\bibliography{references}
\appendix
\section{Derivation of the Dispersion--Dissipation Decomposition of MSE}
\label{appendix:dispersion-dissipation}
In this appendix, we derive the decomposition of the mean squared error (MSE) into dissipation and dispersion components, which is central to the physics-based training loss used in MI2A.
To begin with, we derive \eqref{eqn:mse-BV-improved} starting from the usual definition of the MSE. Let us consider the total MSE ($\tau(t_j )$) at time-step ($t_j$): 
\begin{equation*}
\tau(t_j)
\;=\;
\frac{1}{N}\sum_{i=0}^{N}\bigl(\overline{Y}_{\text{Train},i}^j - X_i^{'j}\bigr)^2,
\end{equation*}
where \(\overline{Y}_{\text{Train}}^j = \bigl\{\overline{Y}_{\text{Train},i}^j \bigr\}_{i=0}^N\) is the ground-truth field at time \(t_j\) and \(X^{'j} = \{X_i^{'j}\}_{i=0}^N\) is the corresponding predicted field.
We next define:
\begin{equation*}
    \langle \overline{Y}_{\text{Train}}^j\rangle 
    := \frac{1}{N}\sum_{i=0}^N \overline{Y}_{\text{Train},i}^j, 
    \quad
    \langle X^{'j}\rangle 
    := \frac{1}{N}\sum_{i=0}^N X_i^{'j},
\end{equation*}
as the spatial means, and 
\begin{equation*}
    \sigma\bigl(\overline{Y}_{\text{Train}}^j\bigr)^2
    := \frac{1}{N}\sum_{i=0}^N 
    \bigl(\overline{Y}_{\text{Train},i}^j - 
    \langle\overline{Y}_{\text{Train}}^j\rangle\bigr)^2, 
    \quad
    \sigma\bigl(X^{'j}\bigr)^2
    := \frac{1}{N}\sum_{i=0}^N 
    \bigl(X_i^{'j} - 
    \langle X^{'j}\rangle\bigr)^2,
\end{equation*}
as the spatial variances (or squared standard deviations). The correlation coefficient \(\rho\) between \(\overline{Y}_{\text{Train}}^j\) and \(X^{'j}\) is
\begin{equation*}
    \rho
    := 
    \frac{\frac{1}{N}
    \sum_{i=0}^N 
    \bigl(\overline{Y}_{\text{Train},i}^j - 
    \langle\overline{Y}_{\text{Train}}^j\rangle\bigr)\,
    \bigl(X_i^{'j} - 
    \langle X^{'j}\rangle\bigr)}
    {\sigma(\overline{Y}_{\text{Train}}^j)\,
     \sigma(X^{'j})}.
\end{equation*}

\subsection*{Step 1: Expand the summation form of the MSE.}
We begin by expanding the squared error term to express $\tau$ in terms of expectations: 
\begin{equation*}
    \tau(t_j)
    = 
    \frac{1}{N}
    \sum_{i=0}^N 
    \Bigl(\overline{Y}_{\text{Train},i}^j - X_i^{'j}\Bigr)^2
    = 
    \frac{1}{N}
    \sum_{i=0}^N 
    \Bigl(\overline{Y}_{\text{Train},i}^j\Bigr)^2
    + 
    \frac{1}{N}
    \sum_{i=0}^N 
    \Bigl(X_i^{'j}\Bigr)^2
    \;-\;
    2\,\frac{1}{N}
    \sum_{i=0}^N 
    \overline{Y}_{\text{Train},i}^j\,X_i^{'j}.
\end{equation*}
Let us denote:
\begin{equation*}
    E\bigl[\overline{Y}_{\text{Train}}^j\bigr] 
    := \langle\overline{Y}_{\text{Train}}^j\rangle, \quad
    E\bigl[X^{'j}\bigr] 
    := \langle X^{'j}\rangle,
\end{equation*}
so that
\begin{equation*}
    E\bigl[\overline{Y}_{\text{Train}}^j\bigr]^2 
    = \langle\overline{Y}_{\text{Train}}^j\rangle^2, \quad
    E\bigl[X^{'j}\bigr]^2 
    = \langle X^{'j}\rangle^2.
\end{equation*}
Using the usual decomposition of variance, we have
\begin{equation*}
    \sigma(\overline{Y}_{\text{Train}}^j)^2
    = 
    E\bigl[\overline{Y}_{\text{Train}}^j{}^2\bigr]
    \;-\;
    E\bigl[\overline{Y}_{\text{Train}}^j\bigr]^2,
    \quad
    \sigma(X^{'j})^2
    = 
    E\bigl[X^{'j}{}^2\bigr]
    \;-\;
    E\bigl[X^{'j}\bigr]^2.
\end{equation*}
Similarly, for the cross-term we get
\begin{equation*}
    E\bigl[\overline{Y}_{\text{Train}}^j\,X^{'j}\bigr]
    = 
    E\bigl[\overline{Y}_{\text{Train}}^j\bigr]\,
    E\bigl[X^{'j}\bigr]
    \;+\;
    \rho\,
    \sigma(\overline{Y}_{\text{Train}}^j)\,
    \sigma(X^{'j}),
\end{equation*}
by definition of the correlation coefficient \(\rho\).

\subsection*{Step 2: Rewriting the MSE}
Putting these forms together and simplifying, we  have:
\begin{equation*}
\begin{aligned}
    \tau(t_j)
    &= 
    \bigl[\,\sigma(\overline{Y}_{\text{Train}}^j)^2
    + 
    \langle\overline{Y}_{\text{Train}}^j\rangle^2\bigr]
    \;+\;
    \bigl[\,\sigma(X^{'j})^2 
    + 
    \langle X^{'j}\rangle^2\bigr]
    \\
    &\qquad
    - 
    2\,
    \bigl[\,
    \langle\overline{Y}_{\text{Train}}^j\rangle\,
    \langle X^{'j}\rangle
    + 
    \rho\,
    \sigma(\overline{Y}_{\text{Train}}^j)\,
    \sigma(X^{'j})\bigr].
\end{aligned}
\end{equation*}
Grouping the mean terms, we can write:
\begin{equation*}
    \langle\overline{Y}_{\text{Train}}^j\rangle^2 
    + 
    \langle X^{'j}\rangle^2 
    - 
    2\,\langle\overline{Y}_{\text{Train}}^j\rangle\,\langle X^{'j}\rangle 
    =
    \Bigl[\langle\overline{Y}_{\text{Train}}^j\rangle - \langle X^{'j}\rangle\Bigr]^2.
\end{equation*}
We now rearrange and group the terms to isolate amplitude-related and correlation-related contributions: 
\begin{equation*}
    \sigma(\overline{Y}_{\text{Train}}^j)^2 
    + 
    \sigma(X^{'j})^2 
    - 
    2\,\rho\,\sigma(\overline{Y}_{\text{Train}}^j)\,\sigma(X^{'j}) 
    \;=\;
    \bigl[\sigma(\overline{Y}_{\text{Train}}^j) - \sigma(X^{'j})\bigr]^2
    \;+\;
    2\,(1-\rho)\,
    \sigma(\overline{Y}_{\text{Train}}^j)\,
    \sigma(X^{'j}).
\end{equation*}
By combining the above expressions, we obtain the total MSE decomposition:
\begin{equation}
\label{eqn:appendix-disp-diss}
    \tau(t_j)
    =
    \bigl[\sigma(\overline{Y}_{\text{Train}}^j) - \sigma(X^{'j})\bigr]^2
    \;+\;
    \bigl(\langle\overline{Y}_{\text{Train}}^j\rangle 
    - 
    \langle X^{'j}\rangle\bigr)^2
    \;+\;
    2\,(1 - \rho)\,
    \sigma(\overline{Y}_{\text{Train}}^j)\,
    \sigma(X^{'j}).
\end{equation}
This result aligns with Eq. \eqref{eqn:mse-BV-improved} presented in the main text and offers a clear decomposition of the MSE, distinguishing between phase and amplitude contributions, which correspond to dispersion and dissipation errors. This decomposition is particularly important for wave propagation tasks, where accurately capturing both the amplitude and phase is essential for long-term prediction. By optimizing the MI2A model using the error decomposition, the network is explicitly guided to correct for both amplitude attenuation and phase distortion.

% \end{linenumbers}

\end{document}